\documentclass[journal]{IEEEtran}
\ifCLASSINFOpdf
\else
\fi

\usepackage[caption=false, font=footnotesize]{subfig}
\usepackage{graphicx,amsmath,verbatim,siunitx,todonotes,changepage}

\newcommand{\reals}{{\mbox{\bf R}}}
\newcommand{\integers}{{\mbox{\bf Z}}}
\newcommand{\symm}{{\mbox{\bf S}}}  

\newcommand{\rank}{\mathop{\bf rank}}

\newcommand{\argmin}{\mathop{\rm argmin}}

\newcommand{\eg}{{\it e.g.}}
\newcommand{\ie}{{\it i.e.}}

\newcommand{\etal}{{\it et al.}}

\newcommand{\normal}{{\mathcal{N}}}

\DeclareSIUnit{\mph}{mph}
\DeclareSIUnit{\ftmin}{ft/min}
\DeclareSIUnit{\feet}{feet}
\DeclareSIUnit{\nm}{NM}

\begin{document}
%
\title{Learning Probabilistic Trajectory Models of Aircraft in Terminal Airspace from Position Data}
%
%
%
\author{Shane~T.~Barratt\IEEEauthorrefmark{1},
        Mykel~J.~Kochenderfer\IEEEauthorrefmark{2},
        and~Stephen~P.~Boyd\IEEEauthorrefmark{1}
\thanks{Manuscript received April 23, 2018; revised August 16, 2018; accepted September 26, 2018. This material is based upon work supported by the National Science Foundation Graduate Research Fellowship under Grant No. DGE-1656518.

All authors are affiliated with Stanford University, Stanford, CA 94305: \IEEEauthorrefmark{1} Department of Electrical Engineering; \IEEEauthorrefmark{2} Department of Aeronautics and Astronautics. E-mails: \texttt{\{sbarratt,mykel,boyd\}@stanford.edu}.

Digital Object Identifier 10.1109/TITS.2018.2877572
}}

%
%

\markboth{IEEE TRANSACTIONS ON INTELLIGENT TRANSPORTATION SYSTEMS}%
{Barratt \etal: Learning Probabilistic Trajectory Models of Aircraft in Terminal Airspace from Position Data}
%



\maketitle

\begin{abstract}
Models for predicting aircraft motion are an important component of modern aeronautical systems.
These models help aircraft plan collision avoidance maneuvers and help conduct offline performance and safety analyses.
In this article, we develop a method for learning a probabilistic generative model of aircraft motion in terminal airspace, the controlled airspace surrounding a given airport.
The method fits the model based on a historical dataset of radar-based position measurements of aircraft landings and takeoffs at that airport.
We find that the model generates realistic trajectories, provides accurate predictions, and captures the statistical properties of aircraft trajectories.
Furthermore, the model trains quickly, is compact, and allows for efficient real-time inference.
\end{abstract}


%
\IEEEpeerreviewmaketitle

\section{Introduction}
\label{sec:introduction}

\begin{figure}[!t]
\centering
\includegraphics[width=.72\columnwidth]{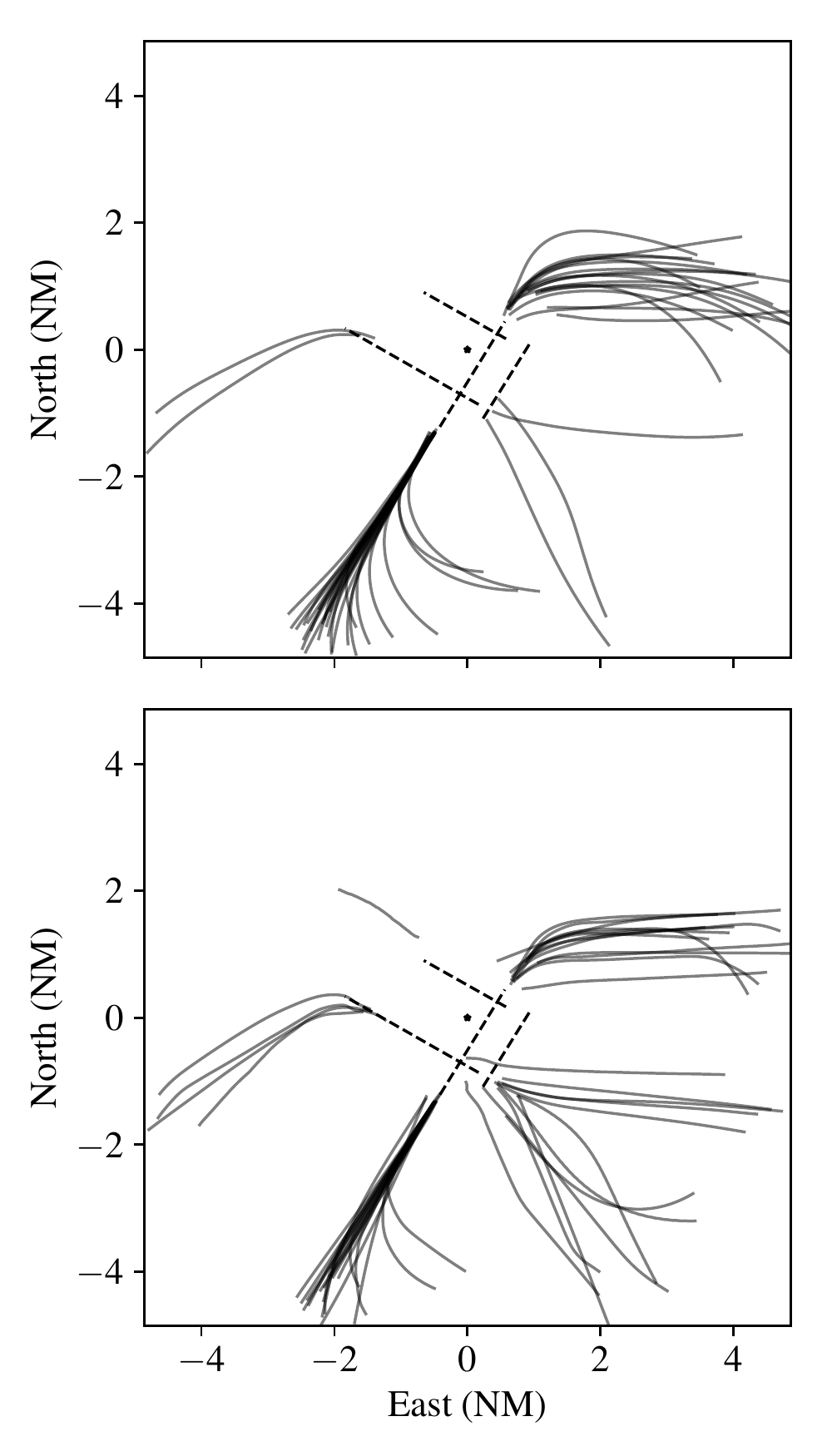}
\caption[]{Lateral view of takeoffs from KJFK.
One figure is of held-out test trajectories and one is of trajectories generated by the model.
Try to guess which one is real and which one is generated before looking at their true identities.\footnotemark}
\label{fig:samplesreal}
\end{figure}
\footnotetext{Top is generated, bottom is held-out test samples.}


\IEEEPARstart{T}{he} United States National Airspace System is undergoing a multibillion dollar modernization to increase safety, efficiency, and predictability of air transport, also known as NextGen~\cite{nextgen}.
A major focus of the modernization is on the portion of flight that occurs in the terminal airspace surrounding an airport, which represents the greatest safety risk.
In fact, \SI{56}{\percent} of fatal air-related accidents worldwide occur in the takeoff, initial climb, final approach, and landing phases of the flight, even though these four phases constitute only \SI{6}{\percent} of flight time~\cite{airplanes1959statistical}.

Probabilistic trajectory models of aircraft in the terminal region are important for the development and assessment of air traffic control technologies.
These trajectory models can be used to predict the future trajectories of aircraft, detect conflicts, and recommend avoidance maneuvers.
In addition, they can be used to assess the safety and operational performance of new technologies and procedures in simulation before deployment.
It is critical, however, that the trajectory models be accurate representations of how aircraft actually behave in the airspace.
This paper focuses on the construction of probabilistic trajectory models of aircraft in terminal airspace from radar-based surveillance data.

{\bf Related work.}
\nocite{enriquez2013identifying}
\nocite{vasquez2004motion}
We begin by summarizing methods for modeling aircraft trajectories in unstructured airspace, \ie, outside of the terminal region.
One line of work involves predicting future aircraft motion by estimating the aircraft's state and propagating that estimate through physical equations of motion~\cite{chatterji1996route,chatterji1999short}.
Other methods include synthesizing trajectories by combining individual segments defined by different modes of operation~\cite{slattery1997trajectory} and estimating vertical paths based on the weight of the aircraft~\cite{warren1998vertical}.
Another method proposed by Kochenderfer \etal~\cite{Kochenderfer2008uncor,kochenderfer2010airspace} is to learn Bayesian network statistical representations of dynamic variables from historical radar data.
Given trajectory predictions and their covariances for two aircraft, Paielli and Erzberger~\cite{paielli1997conflict} derive a method to solve for  \emph{conflict probability}, defined as the probability that the aircraft become too close.
Other methods have been proposed to efficiently calculate conflict probabilities, including probability flow~\cite{van2009fast,pienaar2015application} and importance sampling~\cite{chryssanthacopoulos2010improved}.
Many of these trajectory prediction models are used in production aviation systems, including the Center TRACON Automation System (CTAS)~\cite{coppenbarger1999climb,chan2000improving}, and for evaluating the Traffic Alert and Collision Avoidance System (TCAS)~\cite{henry2000traffic} and ACAS X, the next generation of TCAS~\cite{kochenderfer2012next}.
Realistic airport simulations are also used to validate NextGen~\cite{harkleroad2013risk}.

There are also hybrid dynamics models of aircraft in unstructured airspace, which explicitly take into account both continuous and discrete dynamic behavior.
Prandini \etal~\cite{prandini2000probabilistic} propagate aircraft state through a stochastic hybrid dynamical system model to perform conflict detection.
Combining the stochastic hybrid dynamical system model with a method for inferring the aircraft's navigational intent leads to a commonly used set of modeling tools for aircraft trajectory prediction~\cite{yepes2007new,hwang2008intent,liu2011probabilistic}.
A recently proposed method incorporates the hybrid dynamics model with a Bayesian intent model~\cite{lowe2015learning}.
These methods, for the most part, focus on learning probabilistic trajectory models for aircraft in unstructured airspace.
Modeling aircraft in terminal regions is a fundamentally different task that cannot be solved with models for unstructured operation.
It is difficult to apply such models because the structure of the arrival and departure procedures at airports leads to different behavior.

One way to learn models for the terminal region is by using the airport's published instrument procedures.
However, doing this leads to a model that has no knowledge of the statistical variation in how the procedures are executed.
A way to capture the statistical variation observed in practice is to learn directly from historical surveillance data.

Given the large amount of surveillance data available, several papers have proposed using machine learning techniques to learn models of terminal airspace.
One method involves clustering \emph{turning points}, defined as spatial positions where a substantial change of heading occurs.
Gariel \etal~\cite{gariel2011trajectory} identify and cluster these so-called turning points across a dataset of trajectories, then treat a trajectory as a (discrete) sequence of the cluster labels, then cluster those (discrete) sequences using the least common subsequence algorithm.
Mahboubi and Kochenderfer~\cite{mahboubi2017learning} find that the turning point model does not perform well on real, noisy radar data, and extend the work of Gariel \etal ~by representing the transitions between turning points at small airports using a Gaussian hidden semi-Markov model.

Another promising direction is to cluster trajectories at the level of position measurements, which is also the principle behind air traffic flow modeling of en route traffic~\cite{gariel2010toward,marzuoli2014data}.
The main challenge in clustering this way is that the sequences are of different lengths.
Li~\etal~\cite{li2011anomaly,li2016anomaly} use the density-based spatial clustering of applications with noise (DBSCAN) algorithm to cluster time-aligned trajectories and use the learned clusters for anomaly detection around airports.
They also experiment with clustering the trajectories after performing principal component analysis.
Hong and Lee~\cite{hong2015trajectory} deal with the varying length issue by clustering with the dynamic time warping (DTW) algorithm to discover traffic patterns.
They regress an aircraft's arrival time based on its similarity to the learned clusters and discuss how their method could augment terminal radar approach controllers.
Conde~\etal~\cite{conde2016trajectory} handle the varying length issue by resampling trajectories to make them the same length.
They then cluster the resampled trajectories using the DBSCAN algorithm and construct classifiers to assign new flights to clusters.
Their main application area is resource management.
These three lines of research are the closest to ours.


{\bf Proposed method.}
In this paper, we propose a method for learning probabilistic trajectory models of aircraft in terminal airspaces directly from position measurements.
The model training procedure does not rely on other (likely unavailable) information, \eg, velocity measurements, flight plans, or air traffic control instructions.
Through an unsupervised clustering algorithm, the method is able to discover the departure and approach procedures for an airport, without ever accessing the airport's instrument procedures.
The method then fits a generative model based on the intra-cluster covariance matrices.

The main limitations of the three most similar lines of work are: 1)~they use ad-hoc methods for dealing with the varying length issue; 2)~they do not use the learned clusters to construct a full probabilistic model of motion; and 3)~the DTW and DBSCAN algorithms have trouble scaling to large datasets (our datasets are over $100$ times larger).

Our method deals with the varying length issue, inspired by Li \etal~\cite{li2011anomaly}, by time-aligning trajectories around the time they are closest to the runways and extrapolating the shorter trajectories so that they are all the same length.
This alignment and extrapolation allows us to compute the similarity between trajectrories directly using the Euclidean norm.
Using this similarity metric, we use the K-means clustering algorithm to cluster the trajectories.
Our method then constructs a probabilistic model (a Gaussian mixture model) from the clusters, which leads to accurate inference and realistic generation of trajectories.
The benefits of such an approach are that it is a compact model, trains quickly (which allows for retraining), efficient to sample from and perform inference with, and most importantly easy to understand and ultimately implement.

We find that our method performs well on real, noisy data, and is able to capture relevant statistical properties of the terminal airspace environment.
We also find that trajectories sampled from the model are realistic.
In Fig.~\ref{fig:samplesreal}, we show samples generated from an instance of our model trained on takeoffs at the KJFK airport next to held-out test trajectories.
We evaluate our method throughout the paper on the KJFK airport, for illustration purposes, though applying the method to other airports is trivial.
In fact, another major advantage of our method is that it is relatively automatic; it can produce a generative model from radar data with little to no supervision.
Our method can create generative models for thousands of airports, automatically, from (poor, noisy, and intermittent) radar position data.

{\bf Limitations.}
Our method has several limitations that must be addressed.
It is unable to incorporate the intent of the air traffic controller, is unable to model the complex spatio-temporal interaction between aircraft, and can only fit models based on historical data (and thus cannot be used to test new instrument procedures).


{\bf Summary of contributions:}
\begin{itemize}
\item A linear time trajectory reconstruction procedure given noisy, irregular position measurements.
\item A procedure for automatic unsupervised learning of a Gaussian mixture model for aircraft motion that captures correlation between position dimensions over time.
\item Publicly available code for constructing the model, running the experiments, and pre-trained models for the KJFK airport.\footnote{\texttt{https://github.com/sisl/terminal-airspace-models}}
\end{itemize}

\section{Radar data}
\label{sec:data}

Our method expects a set of aircraft trajectories denoted $\tau_1, \ldots, \tau_{N_\text{traj}}$.
Each trajectory $\tau_i$ is represented by a set of time and position measurement ordered pairs, \ie, $\tau=\{(t_i, \hat p_i)\}_{i=1}^{N_\text{meas}}$ where the time is $t_i \in \integers_+$ and the measurements are $\hat{p}_i \in \reals^3$.

\subsection{FAA radar dataset}

The radar dataset we use in this paper is derived from the Federal Aviation Administration (FAA) multi--sensor fusion tracker~\cite{jagodnik2008fusion}, which fuses radar detections of a single aircraft by multiple sensors into a single track.
The tracker uses primary radars, mode A/C SSRs, mode S SSRs, Wide Area Multilateration systems, and ADS--B receivers.
Each of these surveillance sensor systems has different accuracies, updating rates (\num{1}--\SI{12}{\Hz}), and susceptibility to anomalies.
The data was preprocessed and fused into timestamped entries of target address (used to group entries into flights), position (measured in WGS84 latitude/longitude), pressure altitude (meters) and optionally geometric altitude, lateral velocity, and vertical velocity.
We ignore lateral/vertical velocities in our analysis, as it is unclear whether they are derived from the position measurements or are actual velocity measurements.
The data contains six months of flights starting March 2012 from Central Florida, New York City, and Southern California.

We describe the procedure for filtering the data that corresponds to a particular airport's terminal airspace, defined as the center of its runways $p_\text{ref} \in \reals^3$ in earth--centered, earth--fixed (ECEF) coordinates.
We convert every position measurement to east--north--up (ENU) coordinates determined by the origin $p_\text{ref}$.
We only keep the position measurements that are less than \SI{5}{NM} in the ``east'' or ``north'' dimensions and less than \SI{3000}{\feet} in the ``up'' dimension, \ie, in the terminal airspace of the airport.
Then we iterate through measurements associated with each target address.
Because target addresses are reused, we split sequences of measurements that have a gap in measurements of more than \SI{30}{\second} into separate trajectories.
This filtering results in a set of trajectories specific to a given airport in the format expected by our method, where each trajectory is a set of time and aircraft position measurement pairs.
We shift time so that the first measurement in each trajectory is always at $t=0$.

\subsection{Landing and takeoff separation}
We model landings and takeoffs separately, as these are inherently different modes of operation.
There are many ways to classify a trajectory as a landing or takeoff;
we used the following
heuristic approach.

For each trajectory, let $T=\max_i{t_i}$ (the time of the last measurement in that trajectory), $\text{c}=\argmin_i \|p_i\|_2$ (the index of the measurement closest to the center of the runways), $\dot{z}_{\text{avg}}$ be the average ``up'' velocity in \si{\ftmin}, and $R$ be approximately the length of the longest runway (\SI{2000}{\metre} for KJFK).
The trajectory is then classified as a
\begin{itemize}
\item \emph{landing} if $(\|p_{\text{c}}\|_2 < R) \land (\|p_{T}\|_2 > R) \land (\dot{z}_{\text{avg}} < -\SI{200}{\ftmin} \land (t_c/T > 0.95)$
\item \emph{takeoff} if $(\|p_{\text{c}}\|_2 < R) \land (\|p_0\|_2 > R) \land (\dot{z}_{\text{avg}} > \SI{200}{\ftmin} \land (t_c/T < 0.05)$
\end{itemize}
We used this rule because it seemed to work well with the FAA radar dataset.
Trajectories that do not match either of these rules are discarded.

For landings, we discard the position measurements after the index $c$, making the measurement closest to the origin as the last measurement.
Similarly, for takeoffs, we ignore the position measurements before the index $c$.
For landings, we reverse time in the trajectory by setting $t_i = T - t_i$, making landings and takeoffs both start near the origin.
Reversing time for landings imposes the constraint that for both landings and takeoffs $t_i=0$ is the point that the trajectory is closest to the runways.
From now on, we assume that landings and takeoffs are processed separately.

\subsection{Preprocessing}
The only preprocessing we do is scale each dimension in each measurement by a separate positive scalar.
Pre-scaling allows the dimensions to be compared equally when using $\ell_2$ norms, and is basic standardization practice in machine learning.
For aircraft trajectories, we scale the ``up'' dimension by $10$.
This is a heuristic, guided by the fact that glide slopes are approximately \SIrange{3}{5}{\degree} ($1/\sin{\SI{5}{\degree}}\approx 11$) and by the ratio of the dimensions of the terminal airspace ($\SI{5}{NM}/\SI{3000}{\feet} \approx 10$).
We reverse the scaling after we learn the model.

\section{Model learning}
\label{sec:model}

\begin{figure}[!t]
\centering
\includegraphics[width=\columnwidth]{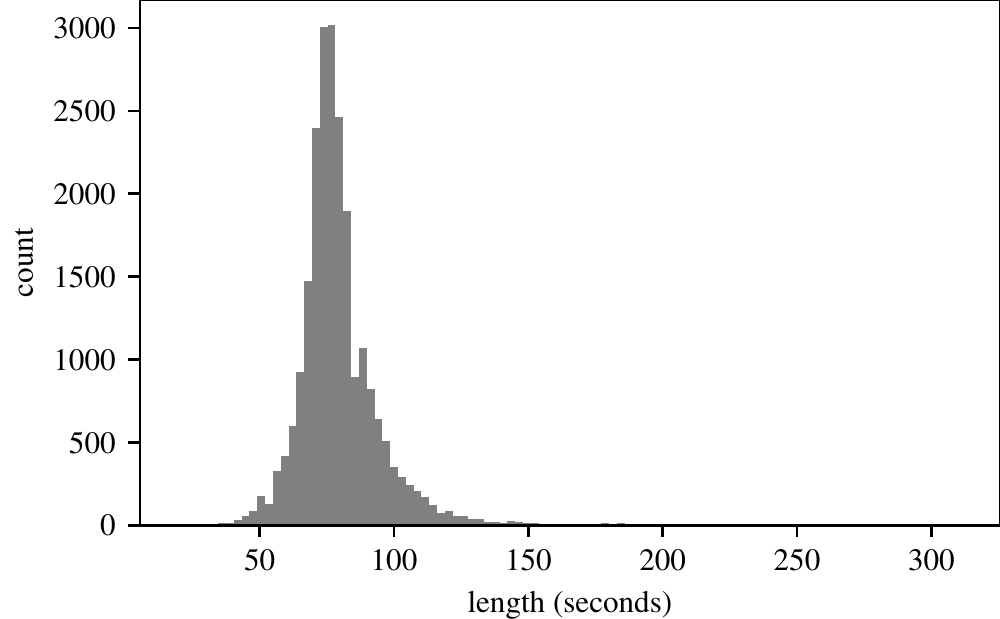}
\caption{Histogram of the length of takeoff trajectories at KJFK in seconds. The median is \num{70} seconds.}
\label{fig:length}
\end{figure}

This section outlines the steps one performs to construct the model.
The steps are to reconstruct the trajectories, cluster them, and then fit a generative model to the clusters.

\subsection{Trajectory reconstruction}
\label{sec:reconstruction}

There are several complications with the trajectories in their original format.
First, the trajectories can have any number (including zero) measurements at a given time step.
Second, the measurements are noisy.
Third, we need to make the trajectories the same duration.

We tackle the third complication by enforcing that the trajectories are a common length, denoted $T_{\text{com}}$ in units of seconds.
We ignore trajectories that are significantly shorter than $T_{\text{com}}$ seconds (\eg, $30$ seconds shorter).
We extrapolate trajectories that are shorter than $T_{\text{com}}$ and truncate trajectories that are longer than $T_{\text{com}}$.

Evidently, the choice of $T_\text{com}$ will have an effect on the model learned,
and should be optimized for the application at hand.
For example, having a larger $T_\text{com}$ leads to less accurate trajectories, as more
trajectories are extrapolated, and a smaller
dataset, as more trajectories are too short.
For illustration, we set $T_\text{com}$ to the median of the trajectory durations across the data (see Fig.~\ref{fig:length} for the histogram of raw trajectory lengths at KJFK).

We interpolate, smooth, and extrapolate by solving the (least-squares) optimization problem
\begin{equation}
\begin{aligned}
& \underset{P}{\text{minimize}}
& & \|AP-\hat{P}\|_F^2 + \lambda_1 \|D_2P\|_F^2 + \lambda_2 \|D_3P\|_F^2,
\end{aligned}
\label{eq:objective}
\end{equation}
where the optimization variable $P \in \reals^{N \times 3}$ is the reconstructed trajectory and its $i$th row is the position of the reconstructed trajectory at time $i$, the length $N=\max\{T, T_{\text{com}}\}$, the diagonal matrix $A \in \reals^{N \times N}$ has $A_{ii}$ equal to one if there are one or more measurements at time $i$ and zero otherwise, each row in the measurement matrix $\hat{P} \in \reals^{N \times 3}$ is the average of the measurements at that time and zero otherwise\footnote{This formulation is equivalent to having an objective function where the $\ell_2$ loss is averaged over each measurement per time step.}, $D_2 \in \reals^{N-2 \times N}$ is the second-order difference matrix representing the acceleration operator, $D_3 \in \reals^{N-4 \times N}$ is the third-order difference matrix representing the jerk operator, $\lambda_1, \lambda_2 \in \reals_+$ are regularization hyper-parameters, and $\|\cdot\|_F$ is the Frobenius norm of a matrix.
We include the form of the (banded) matrices $D_2$ and $D_3$ for the sake of completeness
\begin{equation}
\begin{aligned}
(D_2)_i &= e_i-2e_{i+1}+e_{i+2} \\
(D_3)_i &= -e_{i-1}+2e_i-2e_{i+2}+e_{i+3},
\end{aligned}
\end{equation}
where $(A)_i$ denotes the $i$th row of $A$ and $e_i$ denotes the $i$th standard unit vector.

\begin{figure}[!t]
\centering
\subfloat[]{\includegraphics[width=\columnwidth]{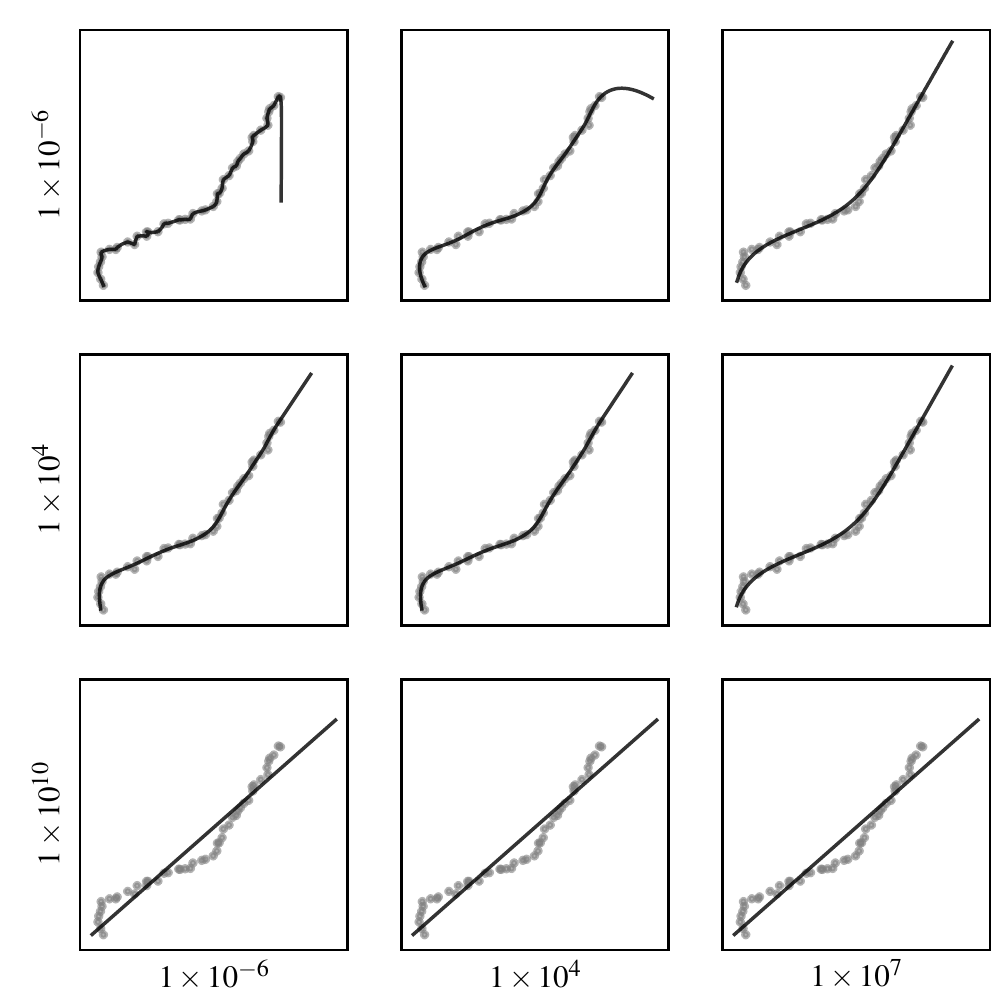}}
\hfill
\subfloat[]{
    \begin{tabular}{cccccccc}
& \multicolumn{1}{c}{} & \multicolumn{3}{c}{$\lambda_2$} \\
&  \multicolumn{1}{c}{} & \multicolumn{1}{c}{$\num{1e-6}$} & \multicolumn{1}{c}{$\num{1e4}$} & \multicolumn{1}{c}{$\num{1e7}$} & & & \\
\cline{3-5}
& $\num{1e-6}$ & \num{1.842} & \num{1.318} & \num{2.590} & & & \\
\cline{3-5}
$\lambda_1$ & \num{1e4} & \num{1.250} & \bf 1.247 & \num{2.590} & & & \\
\cline{3-5}
& $\num{1e10}$ & \num{5.239} & \num{5.239} & \num{5.239}  & & & \\
\cline{3-5}
\end{tabular}
}
\caption{Trajectory reconstruction procedure. (a) Lateral view of measurements and reconstructed trajectory (including a \SI{20}{\second} extrapolation) for various settings of the regularization parameters. A low $\lambda_1$ and $\lambda_2$ leads to a jagged trajectory (top left) whereas high $\lambda_1$ or $\lambda_2$ leads to a linear trajectory (bottom right). (b) Out-of-sample validation losses for various regularization parameters. The optimal out-of-sample validation loss is bolded and achieved by $\lambda_1=\num{1e4}$ and $\lambda_2=\num{1e4}$.}
\label{fig:smoothing}
\end{figure}

The first term of \eqref{eq:objective} encourages the reconstructed trajectory to be close to the measured points and the second and third terms encourage it to have low acceleration and jerk on average.
The regularization parameters $\lambda_1$ and $\lambda_2$ are selected individually for each trajectory through out-of-sample validation.
We perform out-of-sample validation by randomly holding out measurements from that trajectory, fitting trajectories with varying $\lambda_1$ and $\lambda_2$ on the measurements that were not held out, and selecting the parameters that have the lowest loss (\ie, \eqref{eq:objective} with $\lambda_1=\lambda_2=0$) on the held-out measurements.
Fig.~\ref{fig:smoothing} shows the reconstructed trajectories on the Pareto-optimal surface that were generated by varying $\lambda_1$ and $\lambda_2$, as well as the respective out-of-sample losses, for a simulated aircraft trajectory.

The formulation could potentially be improved by switching to a different loss, \eg, Huber or $\ell_1$, and by adding constraints on the reconstructed trajectory~\cite{boyd2004convex}.
We stick to the form in \eqref{eq:objective} mainly for the sake of simplicity of implementation; solving \eqref{eq:objective} reduces to solving a $9$--banded linear system,
which is extremely fast (and linear in $N$).
We believe that the trajectory reconstruction procedure described here could be useful for trajectory smoothing of a wide variety of vehicle data.

\subsection{Clustering}
\label{sec:clustering}

After the trajectories are reconstructed, they all have the same length.
Vectorizing a trajectory (stacking the columns of the matrix $P$) makes it a column vector $p^{(i)} \in \reals^{3T_{\text{com}}}, i = 1, \ldots, N_\text{traj}$.
The first, second, and third parts of the vector are the east, north, and up positions over time.

Since aircraft in terminal airspace, for the most part, follow pre-defined procedures and only slightly deviate from them, it is reasonable to believe that the trajectories partition well into clusters.
Each cluster should have trajectories that follow a similar path.
One way we can measure the similarity between two trajectories $p^{(1)}$ and $p^{(2)}$ is using their Euclidean distance $\|p^{(1)}-p^{(2)}\|_2$.
(One could use other distance metrics, but the Euclidean distance works well with the Gaussian mixture model we eventually fit.)
Given a number of groups, denoted $K$, we want to choose $K$ centers $\mu_j$ and assignments $c_i \in \{1, \ldots, K\}$ to each trajectory to minimize the total distance between each point and its assigned cluster center.
We do this using the K-means++ algorithm~\cite{lloyd1982least,arthur2007k}.

\begin{figure}[t!]
\centering
\includegraphics[width=\columnwidth]{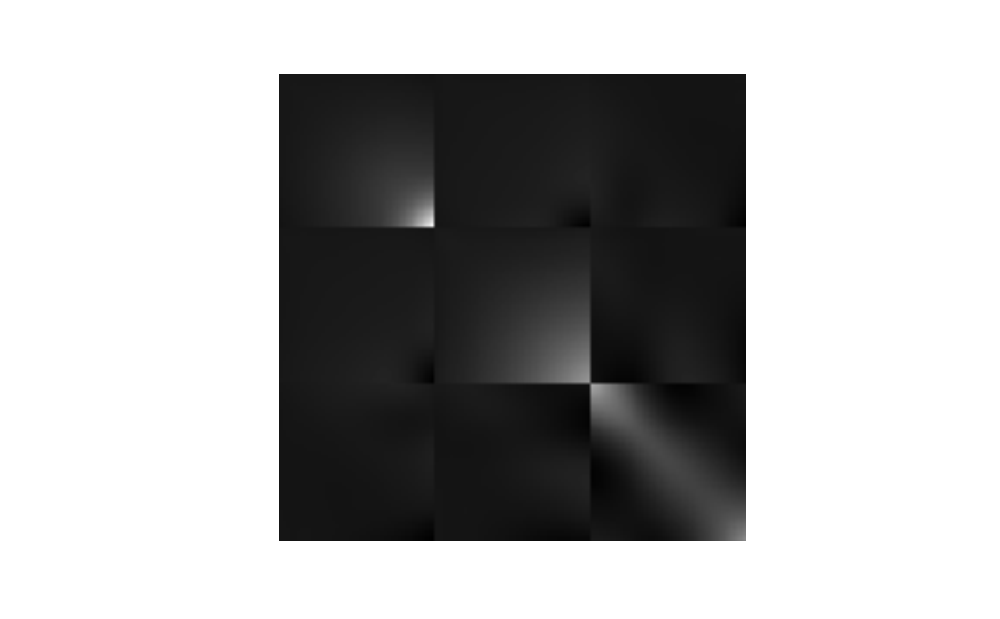}
\caption{Sample covariance matrix for the most probable cluster for takeoffs at KJFK. Higher values of covariance are a lighter shade. The first third of the matrix corresponds to the east dimension, the second third corresponds to north dimension over time, and the final third corresponds to up dimension over time. Uncertainty increases with time, as aircraft are more likely to deviate from their nominal path later in the takeoff maneuver.}
\label{fig:cov}
\end{figure}

The K-means procedure results in $K$ \emph{archetypal trajectories} $\mu_j$ and $N_\text{traj}$ assignments $c_i$.
For each cluster, we calculate the intra-cluster empirical covariance matrices $Q_j \in \symm_{++}^{3T_\text{com}}$.
Fig.~\ref{fig:cov} provides a visualization of one of the intra-cluster covariance matrices.
We let the vector $\pi$ denote cluster frequencies, where $\pi_j$ is the fraction of trajectories in cluster $j$.

\subsection{Gaussian mixture model}
\label{sec:gmm}

Our intra-cluster sample covariance matrices are noisy.
We would like to remove as much of this noise as possible, so we truncate the singular values of the empirical covariance matrix $Q_j$.
(We note that there are more sophisticated methods, \eg,~\cite{friedman2008sparse}.)
The approximation is given by $\tilde{Q}_j = U_j \Sigma_j U_j^T$, where $U_j$ is a matrix with the first $r$ singular vectors, and $\Sigma_j$ is a diagonal matrix with the first $r$ singular values.
The columns of $U_j$ can be interpreted as the \emph{principal deviations} for motion in cluster $j$.
(Note that if we save just the principal deviations instead of $Q_j$ it significantly reduces the overall size of the trained model and speeds up sampling and inference.)
As a final step, we perform the inverse of the preprocessing in Section~\ref{sec:data}, which involves dividing the ``up'' dimension by $10$ and reversing time for landings.

\begin{figure}[!t]
\centering
\includegraphics[width=\columnwidth]{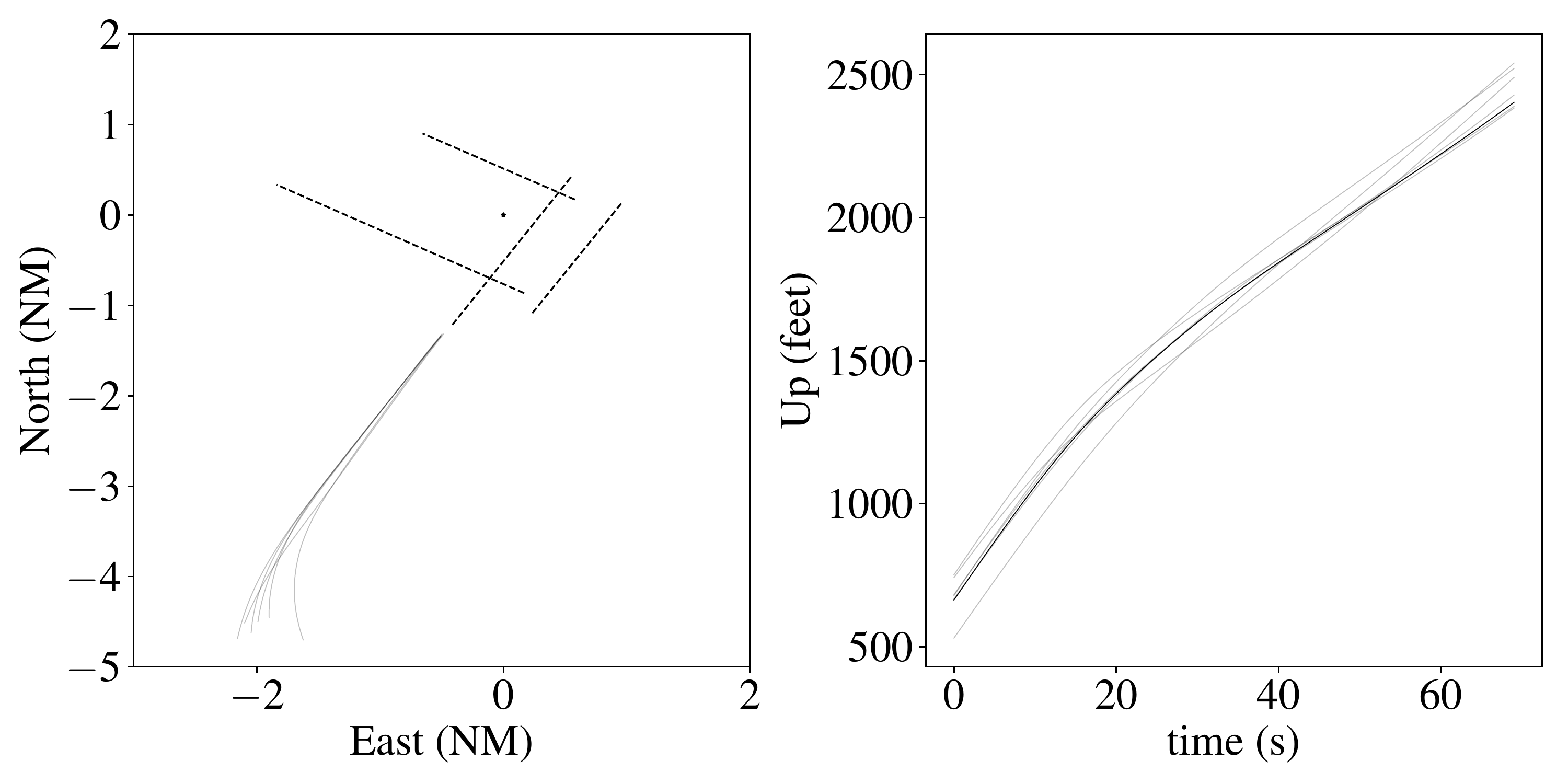}
\caption{Archetypal trajectory (black) and top-5 principal deviations (gray) for the most probable cluster of the model for takeoffs at KJFK with $K=20$.}
\label{fig:principal}
\end{figure}

We keep the top $r=5$ principal deviations for aircraft at KJFK because it captures most of the intra-cluster variance, though in practice $r$ should be chosen using the procedure in Section~\ref{sec:automatic}.
Fig.~\ref{fig:principal} provides a lateral view of the top five principal deviations for the most probable cluster for takeoffs at KJFK.

We now can construct a generative probabilistic model.
Our model for aircraft trajectories in terminal airspaces is the following Gaussian mixture model (GMM):
\begin{equation}
\label{eq:model}
\begin{aligned}
j &\sim \text{Categorical}(\pi) \\
p &\sim \normal(\mu_j, \tilde{Q_j}).
\end{aligned}
\end{equation}
It is a well-known fact that conditioning a Gaussian distribution on measurements gives another Gaussian distribution.
This makes GMMs easy to sample from and perform inference with.


This fact leads to a convenient interpretation of the assumed trajectory distribution.
Because the marginal distribution of a Gaussian is another Gaussian, the first measurement of the trajectory comes from a Gaussian distribution.
Then the position at time $t$ conditioned on the previously sampled points similarly follows a Gaussian distribution.
Drawing a full trajectory from the model is equivalent to successively drawing from Gaussian distributions dependent on the past.

{\bf Positive definiteness of covariance matrices.}
Because of the low-rank approximation step, our covariance matrices are now positive semidefinite and no longer positive definite, which is a requirement to have a valid multivariate density.
We can deal with this inconsistency by restricting the density to a $\rank(Q_j)$ subspace where the Gaussian distribution is supported, which results in the following density:
\begin{equation}
f(x) = (\det{}^*(2\pi Q_j))^{-\frac{1}{2}} \exp(-\frac{1}{2}(x-\mu_j)^T Q_j^{\dagger}(x-\mu_j))\text,
\end{equation}
where $Q_j^{\dagger}$ is the Moore-Penrose inverse of $Q_j$ and $\det{}^*$ is the pseudo-determinant, or the product of the non-zero singular values~\cite{rao1973linear}. 

\begin{figure}[!t]
\centering
\includegraphics[width=.8\columnwidth]{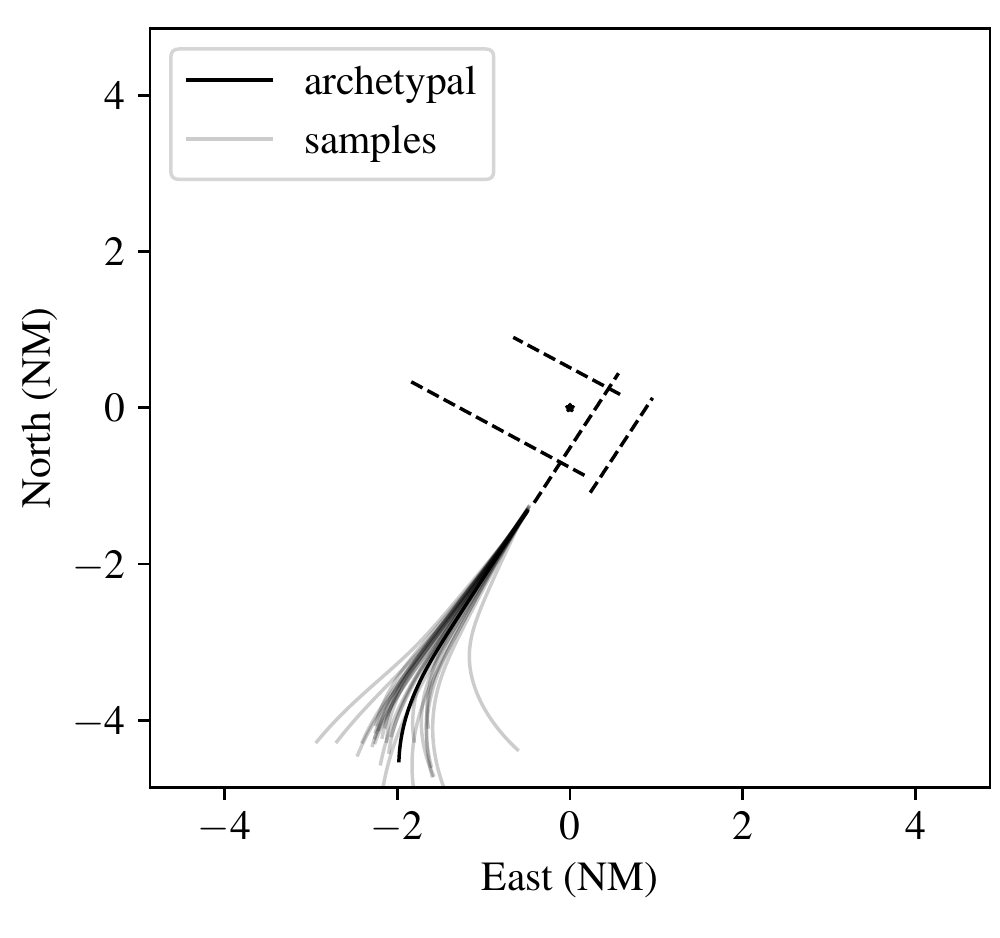}
\caption{Samples from the most probable cluster takeoff model at KJFK with $K=20$. The cluster covariance naturally captures plausible takeoffs at KJFK.}
\label{fig:samples}
\end{figure}

{\bf Sampling.}
Similarly, to sample from cluster $j$ we sample $z \sim \normal(0, I_r)$ and then emit $\mu_j + U_j\Sigma_j^{1/2}z$.
The sample is just an archetypal trajectory plus a random linear combination of the principal deviations weighted by the square root of the singular values, hence the name principal deviations. 
Fig.~\ref{fig:samples} shows a cluster center and samples from that cluster's distribution for the most probable cluster at KJFK.

\subsection{Automatic model generation}
\label{sec:automatic}

We now discuss the claim, made in Section~\ref{sec:introduction}, that the method can go from radar data to generative model with little to no supervision.
The trajectory reconstruction procedure (Section~\ref{sec:reconstruction}) can be performed independently on each trajectory and requires no parameters.
The clustering procedure (Section~\ref{sec:clustering}) requires only one parameter, $K$.
The generative model fitting (Section~\ref{sec:gmm}) also requires only one parameter, $r$.
We can select $K$ and $r$, automatically, using an out-of-sample validation procedure that depends on the downstream task.
First, we split the dataset into a training and held-out set.
We fit generative models with varying values of $K$ and $r$, calculate the performance measure (which depends on the downstream task) of the generative model on the held-out data, and then use the model that achieves the highest performance on the held-out set.
Section~\ref{sec:evaluation} demonstrates this procedure for the tasks of generation and inference.

\section{Experiments}
\label{sec:evaluation}

We evaluate our method on takeoffs and landings at the KJFK airport. Details on the layout of KJFK are given in the Appendixes. 
The plots in the paper are for the KJFK airport with $K=50$ clusters for takeoffs and $K=70$ clusters for landings, as those seemed to be reasonable choices for the number of clusters for the purpose of illustration.
In practice, $K$ should be chosen using the procedure in Section~\ref{sec:automatic}.

\subsection{Archetypal trajectories}

\begin{figure}[!t]
\centering
\includegraphics[width=\columnwidth]{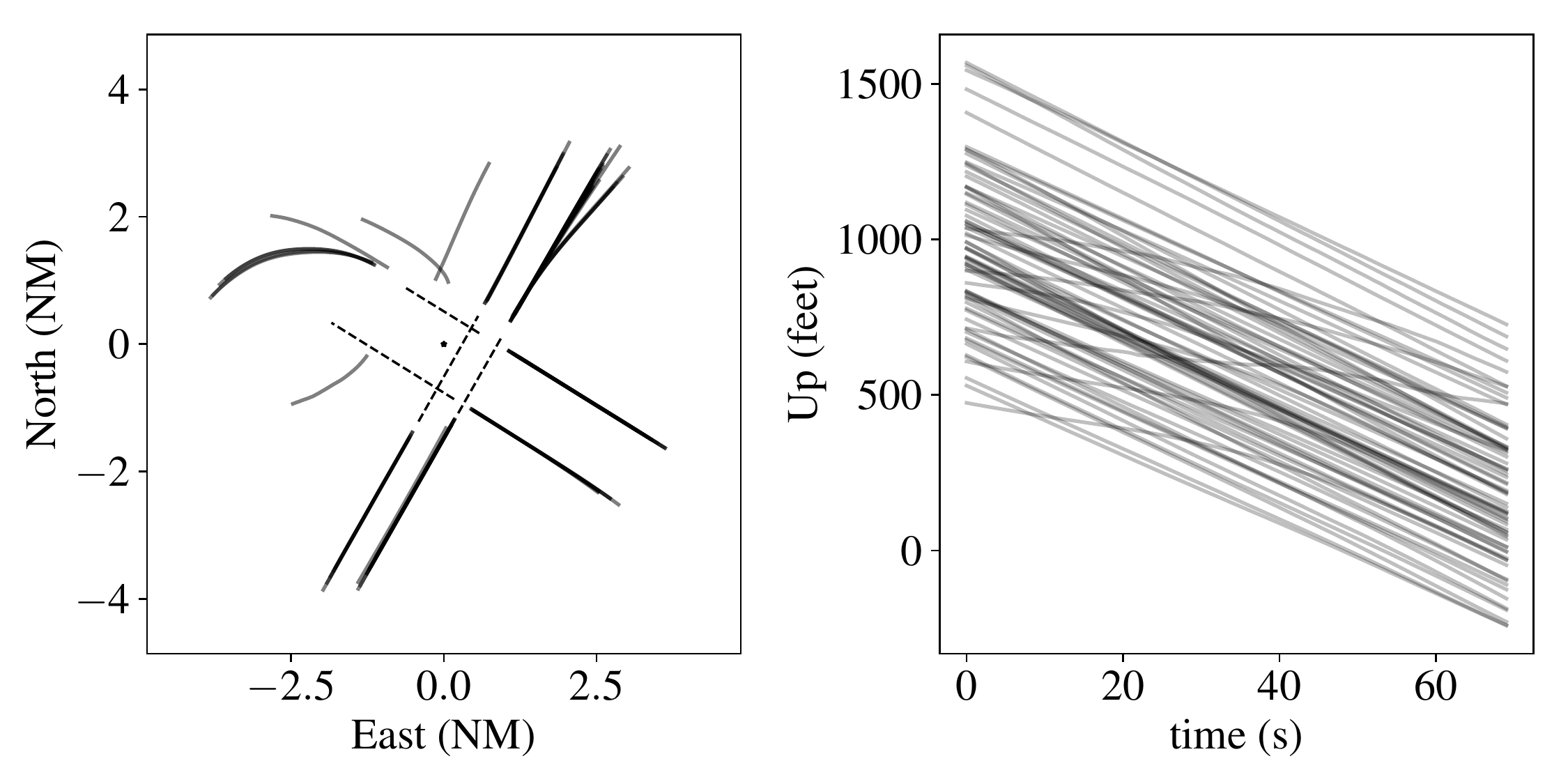}
\caption{Model for landings at KJFK with $K=70$. Left, lateral view of archetypal trajectories. Right, archetypal trajectories vertical profile. Many of the archetypal trajectories overlap in lateral view but not in vertical profile. Each center corresponds to a slightly different glide slope.}
\label{fig:landing-centers}
\end{figure}

\begin{figure}[!t]
\centering
\includegraphics[width=\columnwidth]{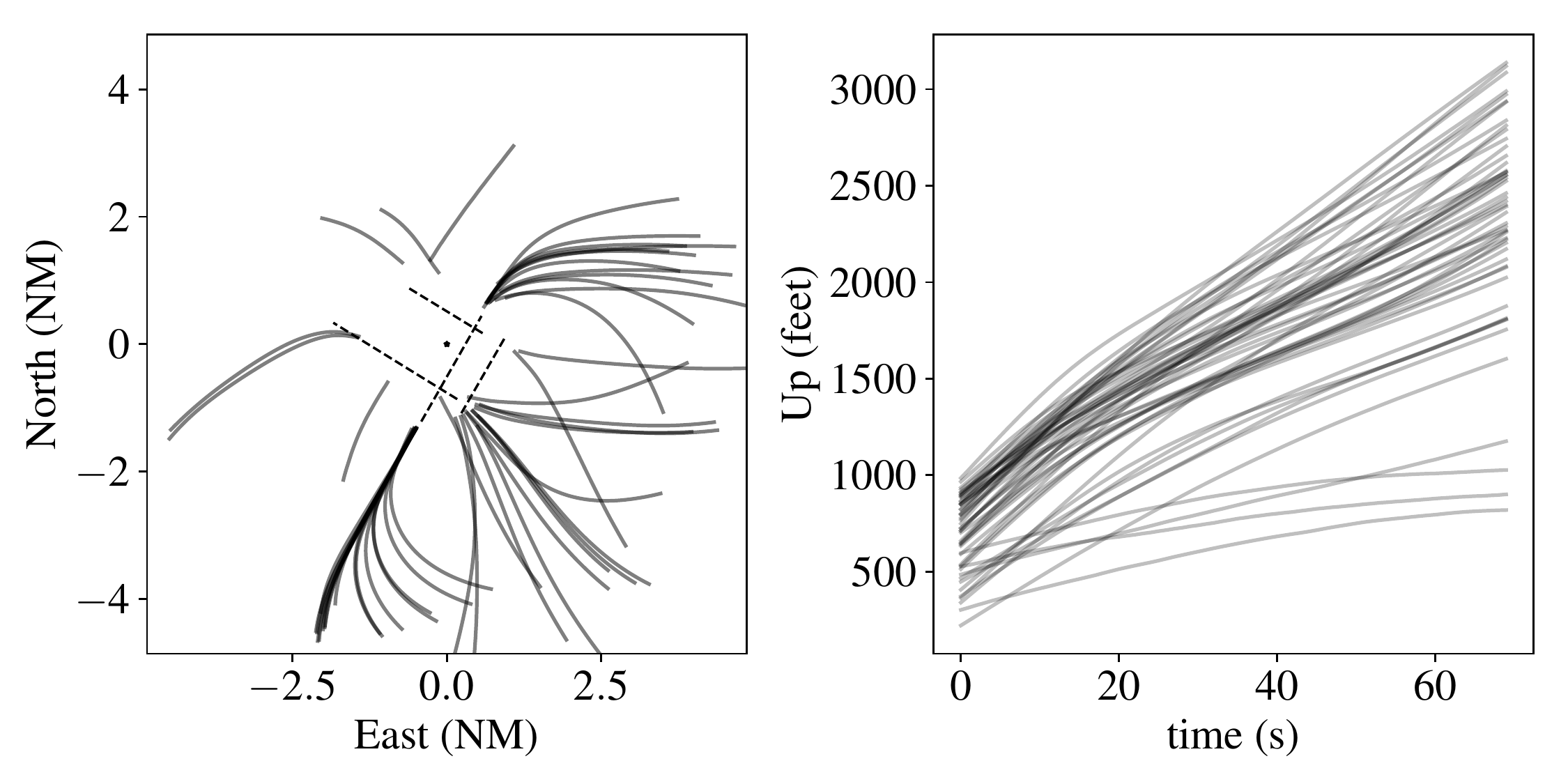}
\caption{Model for takeoffs at KJFK with $K=50$. Left, archetypal trajectories lateral view. Right, archetypal trajectories vertical profile. As with landings, each cluster corresponds to different climb rates.}
\label{fig:takeoff-centers}
\end{figure}

We first sanity check the model by visualizing the learned centers.
Fig.~\ref{fig:landing-centers} shows the learned centers for landings at KJFK and Fig.~\ref{fig:takeoff-centers} shows the centers for takeoffs at KJFK.
The centers seem to capture the plausible ways of approaching and departing KJFK.
Overall, the glide slope for landings seems to be of constant slope and the climb rates for takeoffs seem to start high and then level off.
(The different phases for takeoffs likely correspond to the form of the takeoff procedures at KJFK.)
Several of the archetypal trajectories look super-imposed, which seems redundant.
But in fact, different archetypes can capture not only the procedures, but also approaches at different glide slopes, airspeeds, and wind conditions.
For example, we might end up with clusters for smaller and larger aircraft because they have different characteristics.
These nuances are not coded into the procedure, but rather learned directly from the data.

One observation worth discussing is that there seem to be clusters that represent invalid approaches/departures, \eg, a runway does not exist there or clusters have a negative/constant vertical profile.
These invalid clusters correspond to a small $\pi_j$, and thus low probability, under the model.
That there are clusters representing invalid maneuvers likely means that there are anomalous trajectories in the data. 
These anomalous trajectories could be filtered using outlier detection, but that is out of the scope of this paper.

\subsection{Statistical properties}

To evaluate the statistical properties of the model, we randomly split the trajectories into a training set and heldout set, using a $75$-$25$ training-test split.
Then, we trained the model on the training set and evaluated its performance on the held-out set. To evaluate whether or not the model is able to capture the overall statistics of vehicle motion in the environment, we devised several visual and quantitative experiments and tested the model over different values of $K$.

\begin{figure}[!t]
\centering
\includegraphics[width=\columnwidth]{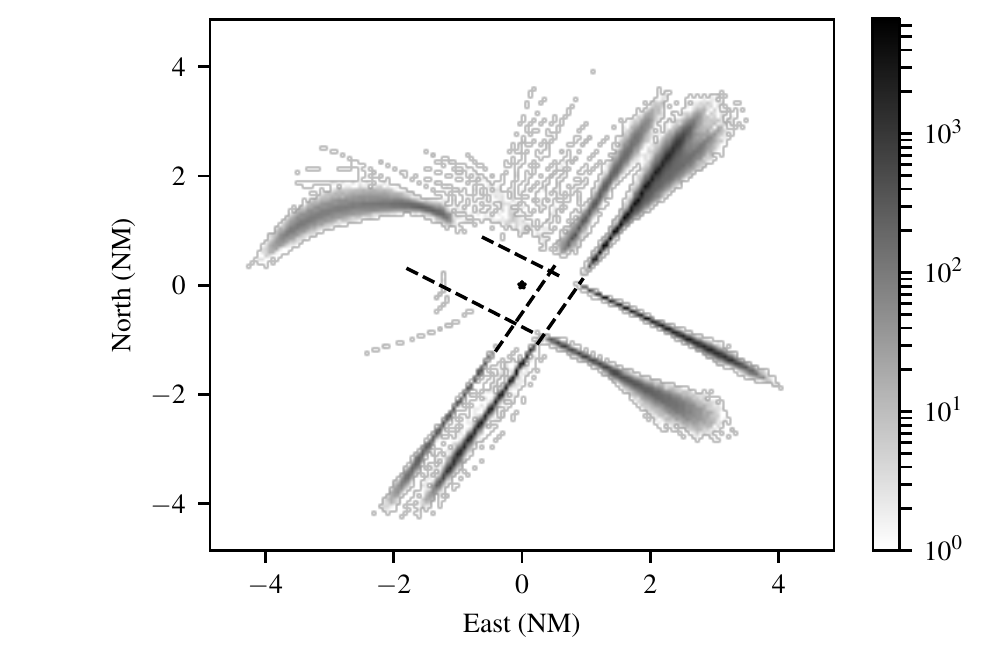}
\caption{Lateral view of log-histogram of {\bf generated} landing flight tracks. Compare with Fig.~\ref{fig:jfk-landings-real}.}
\label{fig:jfk-landings-gen}
\end{figure}

\begin{figure}[!t]
\centering
\includegraphics[width=\columnwidth]{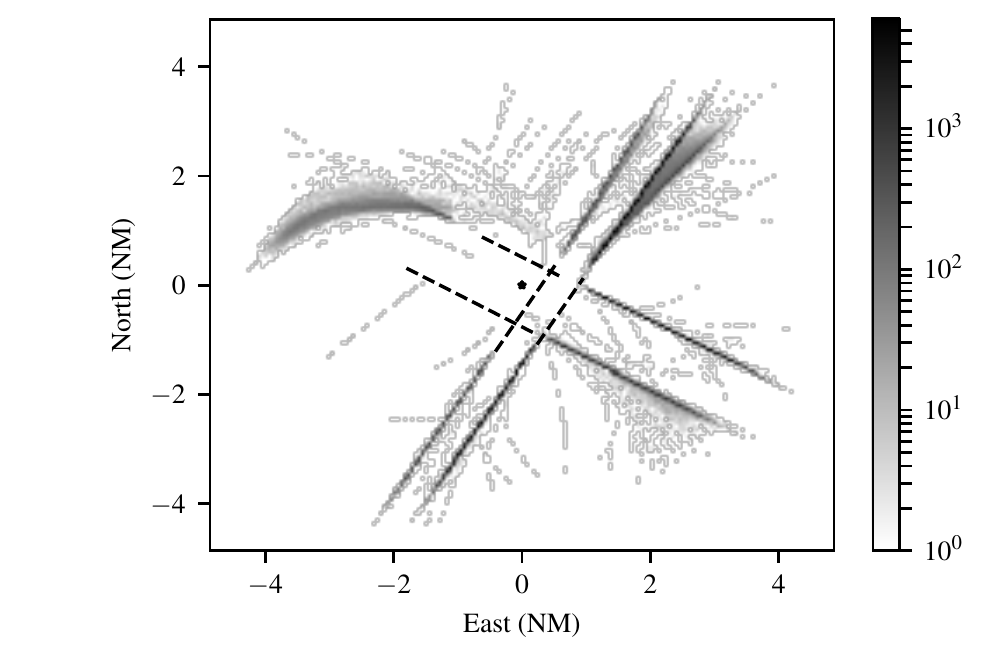}
\caption{Lateral view of log-histogram of {\bf real held-out} landing flight tracks. Compare with Fig.~\ref{fig:jfk-landings-gen}.}
\label{fig:jfk-landings-real}
\end{figure}

\begin{figure}[!t]
\centering
\includegraphics[width=\columnwidth]{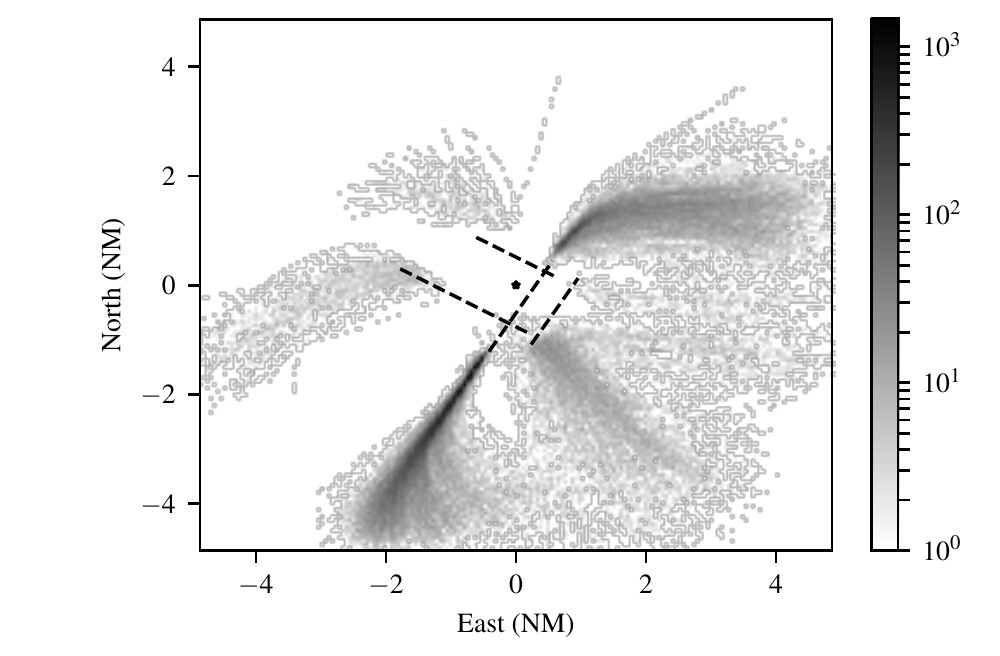}
\caption{Lateral view of log-histogram of {\bf generated} takeoff flight tracks. Compare with Fig.~\ref{fig:jfk-takeoffs-real}.}
\label{fig:jfk-takeoffs-gen}
\end{figure}

\begin{figure}[!t]
\centering
\includegraphics[width=\columnwidth]{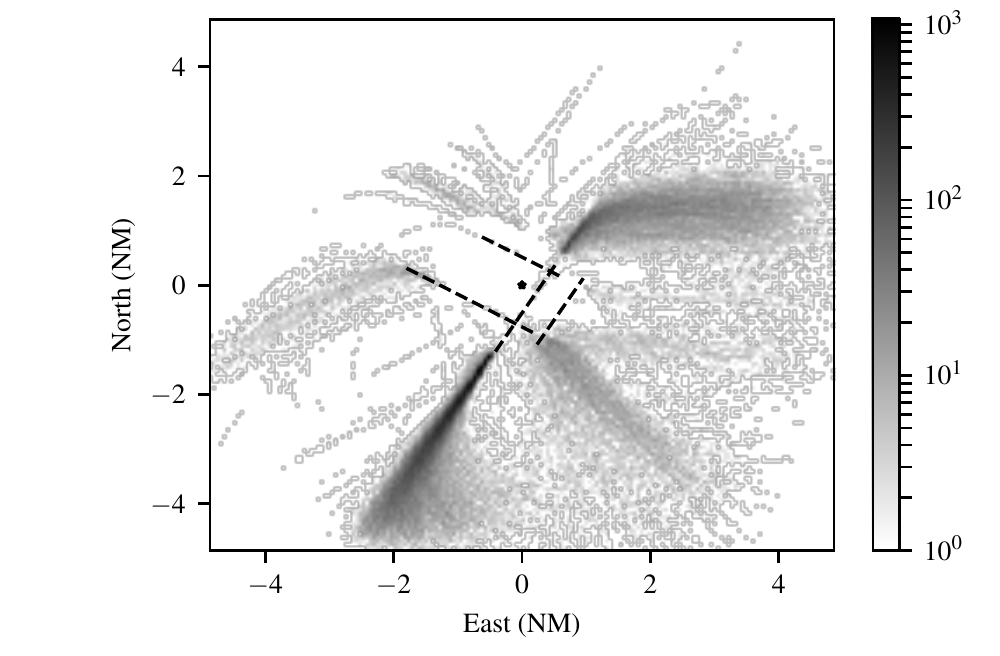}
\caption{Lateral view of log-histogram of {\bf real held-out} takeoff flight tracks. Compare with Fig.~\ref{fig:jfk-takeoffs-gen}.}
\label{fig:jfk-takeoffs-real}
\end{figure}

We have a set of held-out test samples and independently sampled trajectories from the model.
We first plot the overall histogram of positions, with a logarithmic color scale, for generated landings at KJFK (Fig.~\ref{fig:jfk-landings-gen}), held-out landings at KJFK (Fig.~\ref{fig:jfk-landings-real}), generated takeoffs at KJFK (Fig.~\ref{fig:jfk-takeoffs-gen}) and held-out takeoffs at KJFK (Fig.~\ref{fig:jfk-takeoffs-real}).
The $5\times\SI{5}{NM}^2$ grid is split into $400\times400$ bins making each bin $23.15 \times \SI{23.15}{\meter}^2$.

We also would like to make sure that the model captures realistic physical properties of the trajectories outside of just positions, so we calculate the sample longitudinal velocity, vertical velocity, and turn rate for generated samples and compare that to those in held-out data.
Fig.~\ref{fig:lateralhist} in the Appendix shows the generated/held-out turn-rate distributions for takeoffs at KJFK.
(The figures for vertical velocity and turn rate were omitted for the sake of space.)

\begin{figure}[!t]
\centering
\subfloat[]{\includegraphics[width=\columnwidth]{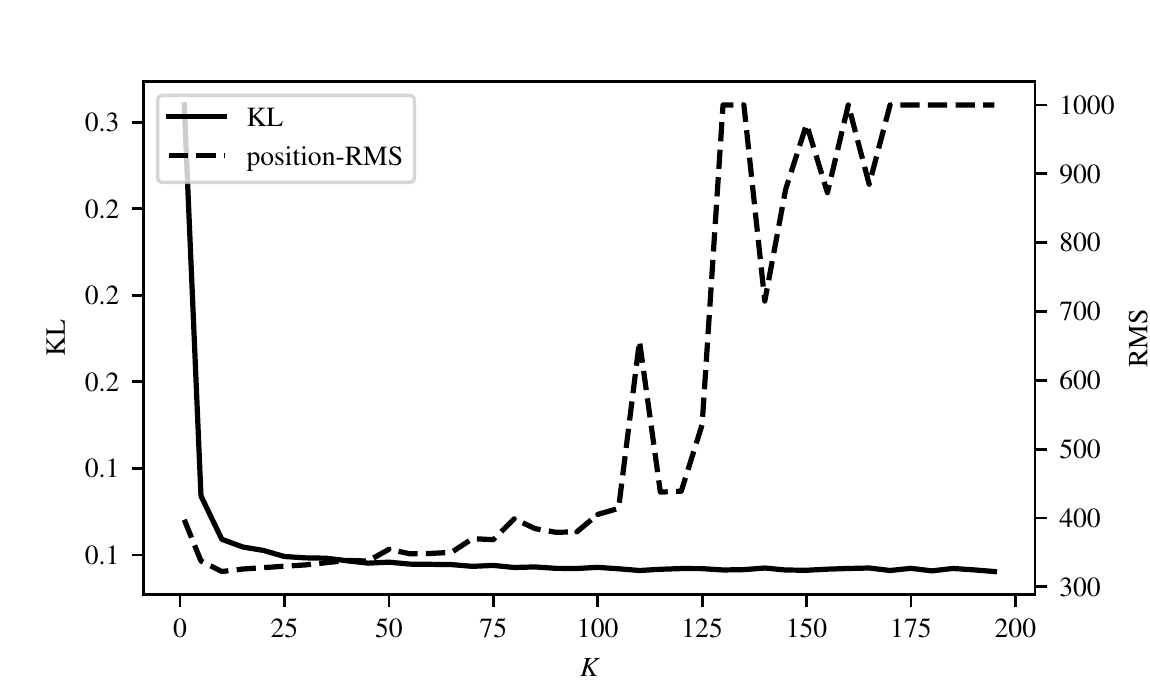}}
\hfill
\subfloat[]{\includegraphics[width=\columnwidth]{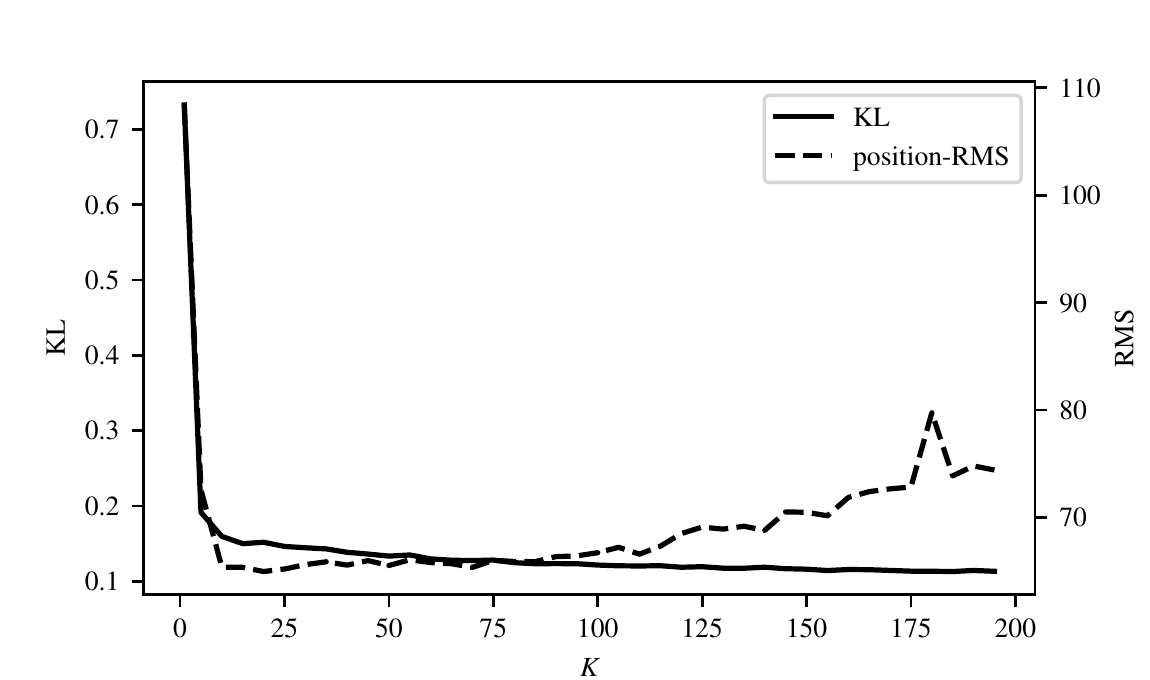}}
\caption{Model for KJFK takeoffs. Log-plot of average KL-divergence and prediction RMS error vs. number of clusters (lower is better in both cases). The RMS error was clipped at \SI{1000}{\meter} for visualization purposes. (a) takeoffs, optimal KL is $0.002$ and the optimal RMS is \SI{322.22}{\meter} ($K=10$); (b) landings, optimal KL is $0.005$ and the optimal RMS is \SI{106.86}{\meter} ($K=20$).}
\label{fig:results}
\end{figure}

Next we performed the out-of-sample validation procedure described in Section~\ref{sec:automatic} for two tasks: generation and prediction.
To evaluate generation, we calculate the Kullback-Leibler divergence (KL-divergence) between the (smoothed) empirical distributions of position, longitudinal velocity, vertical velocity, and turn rate,
\begin{equation}
D_\text{KL} (P \| Q) = - \sum_i P(i) \log \frac{Q(i)}{P(i)}.
\end{equation}
We then average the KL-divergence across the four distributions to get an overall measure of distributional similarity.

To evaluate prediction, we give the model the first $10$ measurements of every held-out trajectory, calculate the closest cluster, and calculate the posterior mean conditioned on those $10$ measurements.
We then calculate the root-mean-square (RMS) error between the posterior mean (the prediction) and the actual rest of the trajectory.
Fig.~\ref{fig:results} shows these two metrics, generation and inference, for varying values of $K$ at KJFK.
(The parameter $r$ was set to $5$ in all experiments.)

For KJFK, from the out-of-sample validation procedure, we can conclude that the optimal number of clusters $K$ for prediction for takeoffs is $10$ (RMS=$322.22$) and for landings is $20$ (RMS=$106.86$).
The inflated RMS value for takeoffs is likely because it is challenging to predict where an aircraft will go given just the direction it took off in, but for landings the model is able to identify the procedure from only $10$ measurements.
Increasing $K$, in the case of prediction, leads to \emph{overfitting}, where the generative model fits the training data too well and fails to generalize to held-out data.
However, for generation, increasing $K$ does not lead to overfitting.
This result is likely because different clusters may represent, \eg, different aircraft types and wind conditions, which are not coded into the airport procedures.
These clusters are not important for RMS-based prediction, but are important for accurately capturing the statistical properties of the terminal airspace environment.

\subsection{Pilot Turing test}

In Fig.~\ref{fig:samplesreal} we showed samples from the model and held-out test samples side-by-side and asked the reader to guess which one was generated and which one was real.
Throughout the experiments, we found the generated samples to be realistic.
To evaluate how realistic the samples are, we devised what we call a ``Pilot Turing Test'' (PTT), akin to the Turing Test in Artificial Intelligence~\cite{turing1948intelligent}.

To perform a PTT, we first take $25$ random samples generated from the model and $25$ random held-out samples.
We added a small amount of random noise to the real and synthetic samples to obscure radar anomalies.
Then, we give the pilot $25$ pairs of generated and real trajectories in random order, where each plot has a lateral view of the trajectory and its vertical profile, as in Fig.~\ref{fig:takeoff-centers}.
We ask the pilot to label one trajectory in each pair as real and one as generated and calculate the accuracy of their predictions.
We want the accuracy to be around \SI{50}{\percent}, which indicates that the generated and real trajectories are indistinguishable.

We conducted a PTT with two licensed pilots for landings/takeoffs at KJFK.
The first pilot (Pilot 1) is a professional Flight Test Captain for a U.S. airline.
The second pilot (Pilot 2) is a professional Boeing 747 pilot with over \num{13000} hours of total flying experience.
The resulting accuracies for both pilots can be found in Table~\ref{tab:trajturing}.
The files used to conduct the PTT can be found in the code.
On average, the pilots achieved an accuracy of $\SI{54}{\percent}$, which means the pilots performed slightly better than random guessing.
We can conclude that our model is able to generate trajectories that are, in practice, visually indistinguishable from real trajectories.

\begin{table}[t]
\centering
\caption{Pilot Turing Test Accuracies}
\label{tab:trajturing}
\begin{tabular}{cccc}
\multicolumn{2}{c}{Pilot 1} & \multicolumn{2}{c}{Pilot 2} \\
\hline
Landings & Takeoffs & Landings & Takeoffs \\
\hline
\rule{0pt}{2ex}
\SI{52}{\percent} & \SI{52}{\percent} & \SI{44}{\percent} & \SI{68}{\percent} \\
\hline
\end{tabular}
\label{ref:results-krg}
\end{table}

\section{Applications}
\label{sec:applications}

This section describes how the generative model can be used to generate trajectories and perform inference.
The model can also be adapted for other applications, including anomaly detection, using the clusters to adjust procedures, and reactionary motion planning.

\subsection{Generation}

As described in Section~\ref{sec:model}, we can draw samples from the assumed probability distribution over aircraft motion.
We draw samples by first sampling from the categorical distribution over clusters, sampling the vector $z \sim \normal(0, I_r)$, and then emitting the trajectory $\mu_j + U_j\Sigma_j^{1/2}z$.
If we further condition the categorical distribution over clusters on external factors such as time of day or previously sampled clusters, we can construct 3D simulations of aircraft operating in particular terminal airspaces.
We could even integrate it into an airport queuing model \cite{idris2002queuing} to construct even more realistic simulations.

The ability to generate samples could be particularly useful for evaluating decision support systems.
For example, the model could be used for verification of AutoATC, an automated air traffic control system for nontowered airports~\cite{mahboubi2015autonomous}.
The model would provide 3D simulations of the terminal airspace, feed the resulting position traces into a simulation of AutoATC, and identify failure cases where AutoATC failed to provide the correct guidance to avoid a collision.
It could also be used to perform ablation studies, \eg, by increasing the ``pressure'' on the airport by increasing the number of clusters sampled per time period or scaling the covariance matrices in the clusters to emit more erratic trajectories.

\subsection{Inference}

\begin{figure}[t!]
\centering
\includegraphics[width=.95\columnwidth]{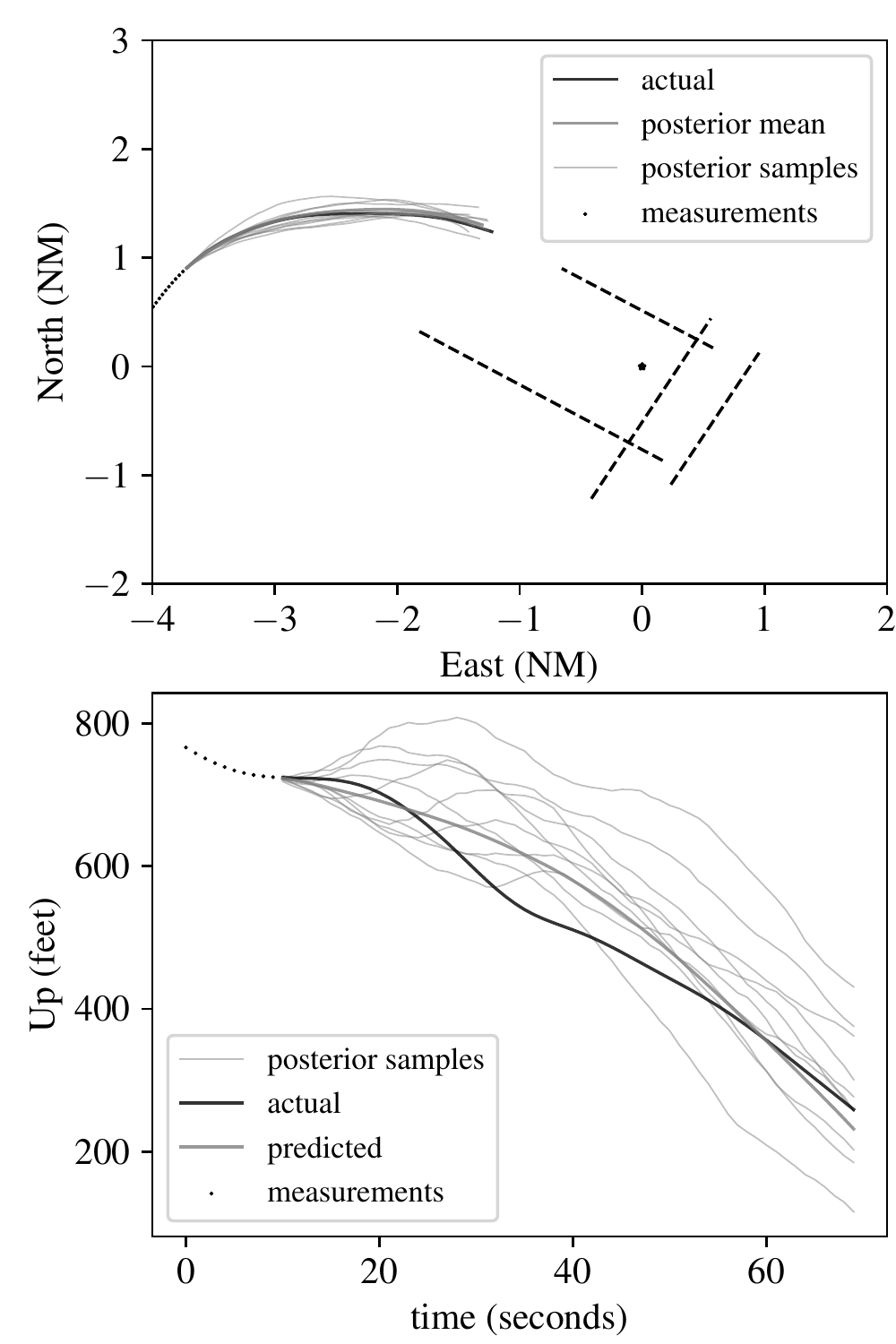}
\caption{Demonstration of inference for a KJFK landing model with $K=20$. Given $10$ samples (dots) of a landing trajectory (black), the model produces a prediction of where the aircraft will go (dark gray) and samples from the posterior distribution (light gray). The average RMS error for the prediction is \SI{77.41}{\meter}.}
\label{fig:inference}
\end{figure}

As described in Section~\ref{sec:model}, we can condition  each cluster's Gaussian distribution on partial measurements.
By performing Bayesian inference in the Gaussian mixture model, we can calculate a probability distribution over clusters and time index in that cluster's distribution.
For each of these possibilities, we can also calculate the posterior distribution over future and past measurements by conditioning that cluster's multivariate Gaussian on the partial measurements.
We demonstrate the result of this procedure for an approach at KJFK in Fig.~\ref{fig:inference}.

Being able to perform inference directly leads to several applications.
If we had, for example, $10$ radar measurements of an aircraft approaching an airport, we can immediately provide non-trivial answers to useful questions (we provide the answers for the inference that was performed in Fig.~\ref{fig:inference}):

\begin{enumerate}
\item Which runway will the aircraft land on? Runway 13L.
\item How long until the aircraft lands? $\approx \SI{60}{\second}$.
\item Which approach procedure is the aircraft performing? PARKWAY VISUAL RWY 13L/R.
\end{enumerate}

If they were departing, we could answer a similar set of questions:

\begin{enumerate}
\item How long until the aircraft exits the terminal airspace?
\item Which direction will the aircraft exit the terminal airspace?
\item Which departure procedure is the aircraft performing?
\end{enumerate}

If we had the posterior distribution for two aircraft, we could perform a Monte Carlo simulation to calculate a conflict probability by sampling a large number of possible trajectories from both posterior distributions and calculating the proportion that violated some conflict criteria.
This posterior distribution could also be used as a model of other aircraft in an on-board planning system for collision avoidance, or as input to an automated air traffic control system.

\section{Conclusion}
\label{sec:conclusion}

In this paper, we proposed a method for learning probabilistic motion models for aircraft in terminal airspaces directly from position measurements.
Our method could be extended to a larger terminal airspace, \eg, $<\SI{50}{NM}$ from the airport, and be used to aid terminal radar approach controllers as in Hong and Lee~\cite{hong2015trajectory}.
We also believe that a slightly altered version of our method could be applied to vehicles in other contexts.
It is likely to work well for highly structured environments, \eg, loading dock for trucks, parking lots, intersections with stoplights/stop signs, or shipping ports.
On the other hand, the method will not perform as well in unstructured environments, such as interstate freeways and in unstructured airspace.

The main limitation and hence direction for future work is to incorporate pairwise interaction.
We can assume that aircraft actively avoid each other, which affects the probability distribution over future motion.
We believe that modeling pairwise interaction can be achieved with a framework similar to ours.

\appendices

\section{KJFK airport}
\label{sec:airports}

The John F Kennedy International Airport (KJFK) airport is a towered airport in Queens, New York with four runways.
As of 2016, KJFK has 1256 aircraft operations per day, \SI{91}{\percent} commercial, \SI{7}{\percent} air taxi, \SI{2}{\percent} transient general aviation and less than \SI{1}{\percent} military~\cite{airnavkjfk}.



\section{Supplemental Figures}
\label{sec:plots}
\begin{figure}[!ht]
\centering
\includegraphics[width=\columnwidth]{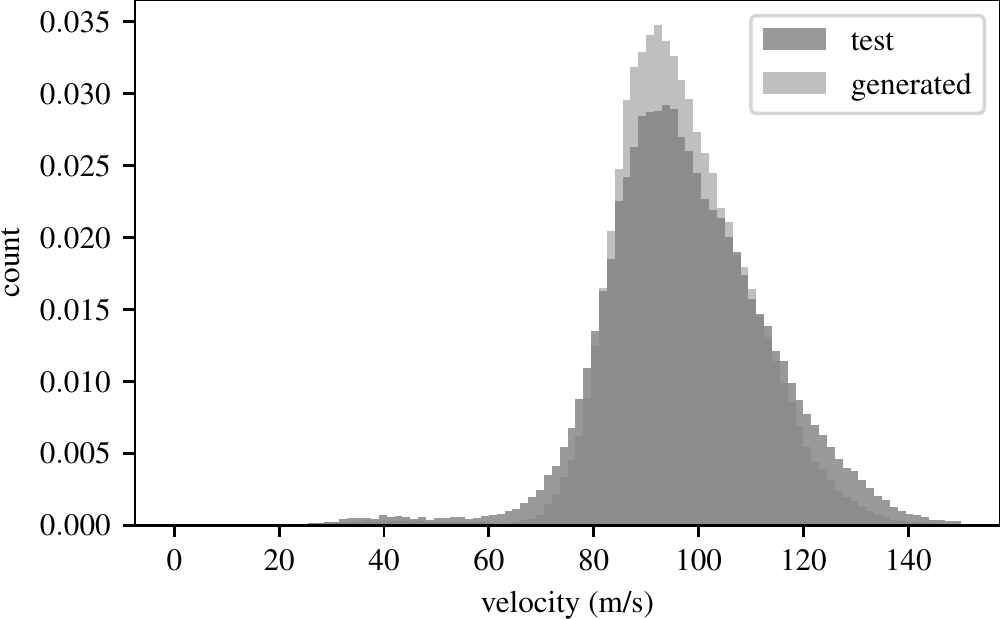}
\caption{Model for takeoffs at KJFK with $K=50$. Histogram of lateral velocities for generated and held-out test data.}
\label{fig:lateralhist}
\vskip -2em
\end{figure}
\section*{Acknowledgments}

The authors acknowledge the FAA and the MITRE Corporation for sharing the surveillance data.
The authors also thank E. Maki for thoughtful discussion, and R. Abraham and G. Knox for their participation in the Pilot Turing Test.

\ifCLASSOPTIONcaptionsoff
  \newpage
\fi



\bibliographystyle{IEEEtran}
\bibliography{bibtex/bib/IEEEexample}

\begin{thebibliography}{10}
\providecommand{\url}[1]{#1}
\csname url@samestyle\endcsname
\providecommand{\newblock}{\relax}
\providecommand{\bibinfo}[2]{#2}
\providecommand{\BIBentrySTDinterwordspacing}{\spaceskip=0pt\relax}
\providecommand{\BIBentryALTinterwordstretchfactor}{4}
\providecommand{\BIBentryALTinterwordspacing}{\spaceskip=\fontdimen2\font plus
\BIBentryALTinterwordstretchfactor\fontdimen3\font minus
  \fontdimen4\font\relax}
\providecommand{\BIBforeignlanguage}[2]{{%
\expandafter\ifx\csname l@#1\endcsname\relax
\typeout{** WARNING: IEEEtran.bst: No hyphenation pattern has been}%
\typeout{** loaded for the language `#1'. Using the pattern for}%
\typeout{** the default language instead.}%
\else
\language=\csname l@#1\endcsname
\fi
#2}}
\providecommand{\BIBdecl}{\relax}
\BIBdecl

\bibitem{nextgen}
``{Modernization of U.S. Airspace},'' https://www.faa.gov/nextgen, 2018.

\bibitem{airplanes1959statistical}
Boeing, ``Statistical summary of commercial jet airplane accidents,'' Boeing,
  Tech. Rep., 2016.

\bibitem{enriquez2013identifying}
M.~Enriquez, ``Identifying temporally persistent flows in the terminal airspace
  via spectral clustering,'' in \emph{USA/Europe Air Traffic Management
  Research and Development Seminar}, 2013.

\bibitem{vasquez2004motion}
D.~Vasquez and T.~Fraichard, ``Motion prediction for moving objects: a
  statistical approach,'' in \emph{Intl. Conf. Robotics Automation},
  vol.~4.\hskip 1em plus 0.5em minus 0.4em\relax IEEE, 2004, pp. 3931--3936.

\bibitem{chatterji1996route}
G.~Chatterji, B.~Sridhar, and K.~Bilimoria, ``En-route flight trajectory
  prediction for conflict avoidance and traffic management,'' in \emph{Proc.
  AIAA Guidance, Navigation, and Control Conf.}, July 1996, p. 3766.

\bibitem{chatterji1999short}
G.~Chatterji, ``Short-term trajectory prediction methods,'' in \emph{Proc. AIAA
  Guidance, Navigation, and Control Conf.}, 1999, p. 4233.

\bibitem{slattery1997trajectory}
R.~Slattery and Y.~Zhao, ``Trajectory synthesis for air traffic automation,''
  \emph{Journal Guidance, Control Dynamics}, vol.~20, no.~2, pp. 232--238,
  1997.

\bibitem{warren1998vertical}
A.~Warren and Y.~Ebrahimi, ``Vertical path trajectory prediction for next
  generation atm,'' in \emph{Proc. 17th Digital Avionics Systems Conf.},
  vol.~2.\hskip 1em plus 0.5em minus 0.4em\relax IEEE, 1998, pp. F11--1.

\bibitem{Kochenderfer2008uncor}
M.~Kochenderfer, J.~Kuchar, L.~Espindle, and J.~Griffith, ``Uncorrelated
  encounter model of the national airspace system,'' Massachusetts Institute of
  Technology, Lincoln Laboratory, Project Report ATC-345, 2008.

\bibitem{kochenderfer2010airspace}
M.~Kochenderfer, M.~Edwards, L.~Espindle, J.~Kuchar, and D.~Griffith,
  ``Airspace encounter models for estimating collision risk,'' \emph{Journal
  Guidance, Control, Dynamics}, vol.~33, no.~2, pp. 487--499, 2010.

\bibitem{paielli1997conflict}
R.~Paielli and H.~Erzberger, ``Conflict probability estimation for free
  flight,'' \emph{Journal Guidance, Control Dynamics}, vol.~20, no.~3, pp.
  588--596, Oct 1997.

\bibitem{van2009fast}
C.~van Daalen and T.~Jones, ``Fast conflict detection using probability flow,''
  in \emph{Automatica}, vol.~45, no.~8, Aug 2009, pp. 1903--1909.

\bibitem{pienaar2015application}
L.~Pienaar and T.~Jones, ``The application of probability flow for conflict
  detection near airports,'' in \emph{Proc. AIAA Guidance, Navigation, and
  Control Conf.}, Jan 2015, p. 1325.

\bibitem{chryssanthacopoulos2010improved}
J.~Chryssanthacopoulos, M.~Kochenderfer, and R.~Williams, ``Improved {Monte
  Carlo} sampling for conflict probability estimation,'' in \emph{Proc. AIAA
  Structures, Structural Dynamics, Materials Conf.}, Apr 2010, p. 3012.

\bibitem{coppenbarger1999climb}
R.~Coppenbarger, ``Climb trajectory prediction enhancement using airline
  flight-planning information,'' in \emph{Proc. AIAA Guidance, Navigation, and
  Control Conf.}, 1999, p. 170.

\bibitem{chan2000improving}
W.~Chan, R.~Bach, and J.~Walton, ``Improving and validating ctas performance
  models,'' in \emph{Proc. AIAA Guidance, Navigation, and Control Conf.}, Aug
  2000, p. 4476.

\bibitem{henry2000traffic}
RTCA, ``{Minimum Operational Performance Standards for Traffic Alert and
  Collision Avoidance System II (TCAS II)},'' RTCA DO-185B, Washington, D.C.,
  Tech. Rep., June 2008.

\bibitem{kochenderfer2012next}
M.~Kochenderfer, J.~Holland, and J.~Chryssanthacopoulos, ``Next generation
  airborne collision avoidance system,'' \emph{Lincoln Laboratory Journal},
  vol.~19, no.~1, pp. 17--33, 2012.

\bibitem{harkleroad2013risk}
E.~Harkleroad, A.~Vela, J.~Kuchar, B.~Barnett, and R.~Merchant-Bennett,
  ``Risk-based modeling to support nextgen concept assessment and validation,''
  \emph{Lincoln Lab Report, ATC-405}, 2013.

\bibitem{prandini2000probabilistic}
M.~Prandini, J.~Hu, J.~Lygeros, and S.~Sastry, ``A probabilistic approach to
  aircraft conflict detection,'' \emph{IEEE Trans. Intell. Transp. Syst.},
  vol.~1, no.~4, pp. 199--220, Dec 2000.

\bibitem{yepes2007new}
J.~Yepes, I.~Hwang, and M.~Rotea, ``New algorithms for aircraft intent
  inference and trajectory prediction,'' \emph{Journal Guidance, Control
  Dynamics}, vol.~30, no.~2, pp. 370--382, 2007.

\bibitem{hwang2008intent}
I.~Hwang and C.~Seah, ``{Intent-Based} probabilistic conflict detection for the
  next generation air transportation system,'' in \emph{Proc. IEEE}, vol.~96,
  no.~12, Dec 2008, pp. 2040--2059.

\bibitem{liu2011probabilistic}
W.~Liu and I.~Hwang, ``Probabilistic trajectory prediction and conflict
  detection for air traffic control,'' \emph{Journal Guidance, Control
  Dynamics}, vol.~34, no.~6, pp. 1779--1789, 2011.

\bibitem{lowe2015learning}
C.~Lowe and J.~How, ``Learning and predicting pilot behavior in uncontrolled
  airspace,'' in \emph{Proc. AIAA Infotech at Aerospace}, 2015, p. 1199.

\bibitem{gariel2011trajectory}
M.~Gariel, A.~Srivastava, and E.~Feron, ``Trajectory clustering and an
  application to airspace monitoring,'' \emph{IEEE Trans. Intell. Transp.
  Syst.}, vol.~12, no.~4, pp. 1511--1524, July 2011.

\bibitem{mahboubi2017learning}
Z.~Mahboubi and M.~Kochenderfer, ``Learning traffic patterns at small airports
  from flight tracks,'' \emph{IEEE Trans. Intell. Transp. Syst.}, vol.~18,
  no.~4, pp. 917--926, 2017.

\bibitem{gariel2010toward}
M.~Gariel, ``Toward a graceful degradation of air traffic management systems,''
  Ph.D. dissertation, Georgia Institute of Technology, 2010.

\bibitem{marzuoli2014data}
A.~Marzuoli, M.~Gariel, A.~Vela, and E.~Feron, ``Data-based modeling and
  optimization of en route traffic,'' \emph{Journal Guidance, Control,
  Dynamics}, vol.~37, no.~6, pp. 1930--1945, 2014.

\bibitem{li2011anomaly}
L.~Li, M.~Gariel, R.~Hansman, and R.~Palacios, ``Anomaly detection in
  onboard-recorded flight data using cluster analysis,'' in \emph{Proc. 2011
  IEEE/AIAA 30th Digital Avionics Systems Conf.}, Oct 2011, pp. 4A4--1.

\bibitem{li2016anomaly}
L.~Li, J.~Hansman, R.~Palacios, and R.~Welsch, ``Anomaly detection via a
  gaussian mixture model for flight operation and safety monitoring,''
  \emph{Transportation Research Part C: Emerging Technologies}, vol.~64, pp.
  45--57, 2016.

\bibitem{hong2015trajectory}
S.~Hong and K.~Lee, ``Trajectory prediction for vectored area navigation
  arrivals,'' \emph{Journal of Aerospace Information Systems}, vol.~12, no.~7,
  pp. 490--502, 2015.

\bibitem{conde2016trajectory}
M.~Conde Rocha~Murca, R.~DeLaura, J.~Hansman, R.~Jordan, T.~Reynolds, and
  H.~Balakrishnan, ``Trajectory clustering and classification for
  characterization of air traffic flows,'' in \emph{Proc. 16th AIAA Aviation
  Technology, Integration, and Operations Conf.}, 2016, p. 3760.

\bibitem{jagodnik2008fusion}
C.~Jagodnik, J.~Stella, and D.~Varon, ``Fusion tracking in air traffic
  control,'' \emph{Journal Air Traffic Control}, vol.~50, no.~1, May 2008.

\bibitem{boyd2004convex}
S.~Boyd and L.~Vandenberghe, \emph{Convex optimization}.\hskip 1em plus 0.5em
  minus 0.4em\relax Cambridge University Press, 2004.

\bibitem{lloyd1982least}
S.~Lloyd, ``Least squares quantization in {PCM},'' \emph{IEEE Trans. Inf.
  Theory}, vol.~28, no.~2, pp. 129--137, Mar 1982.

\bibitem{arthur2007k}
D.~Arthur and S.~Vassilvitskii, ``k-means++: The advantages of careful
  seeding,'' in \emph{Proc. ACM-SIAM Symp. on Discrete Algorithms}, Jan 2007,
  pp. 1027--1035.

\bibitem{friedman2008sparse}
J.~Friedman, T.~Hastie, and R.~Tibshirani, ``Sparse inverse covariance
  estimation with the graphical lasso,'' \emph{Biostatistics}, vol.~9, no.~3,
  pp. 432--441, 2008.

\bibitem{rao1973linear}
C.~Rao, \emph{Linear Statistical Inference and its Applications}.\hskip 1em
  plus 0.5em minus 0.4em\relax Wiley New York, 1973, vol.~2.

\bibitem{turing1948intelligent}
A.~Turing, ``Intelligent machinery, a heretical theory,'' \emph{Philosophia
  Mathematica}, vol.~4, pp. 256--260, 1996.

\bibitem{idris2002queuing}
H.~Idris, J.~Clarke, R.~Bhuva, and L.~Kang, ``Queuing model for taxi-out time
  estimation,'' \emph{Air Traffic Control Quarterly}, vol.~10, no.~1, pp.
  1--22, 2002.

\bibitem{mahboubi2015autonomous}
Z.~Mahboubi and M.~Kochenderfer, ``Autonomous air traffic control for
  non-towered airports,'' in \emph{Proc. USA/Europe Air Traffic Manage. Res.
  and Dev. Seminar}, 2015, pp. 1--6.

\bibitem{airnavkjfk}
``{AirNav: KJFK - John F Kennedy Int. Airport},''
  http://www.airnav.com/airport/KJFK, 2018.

\end{thebibliography}
\vskip -3.25\baselineskip 
\begin{IEEEbiography}[{\includegraphics[width=1in,height=1.25in,clip,keepaspectratio]{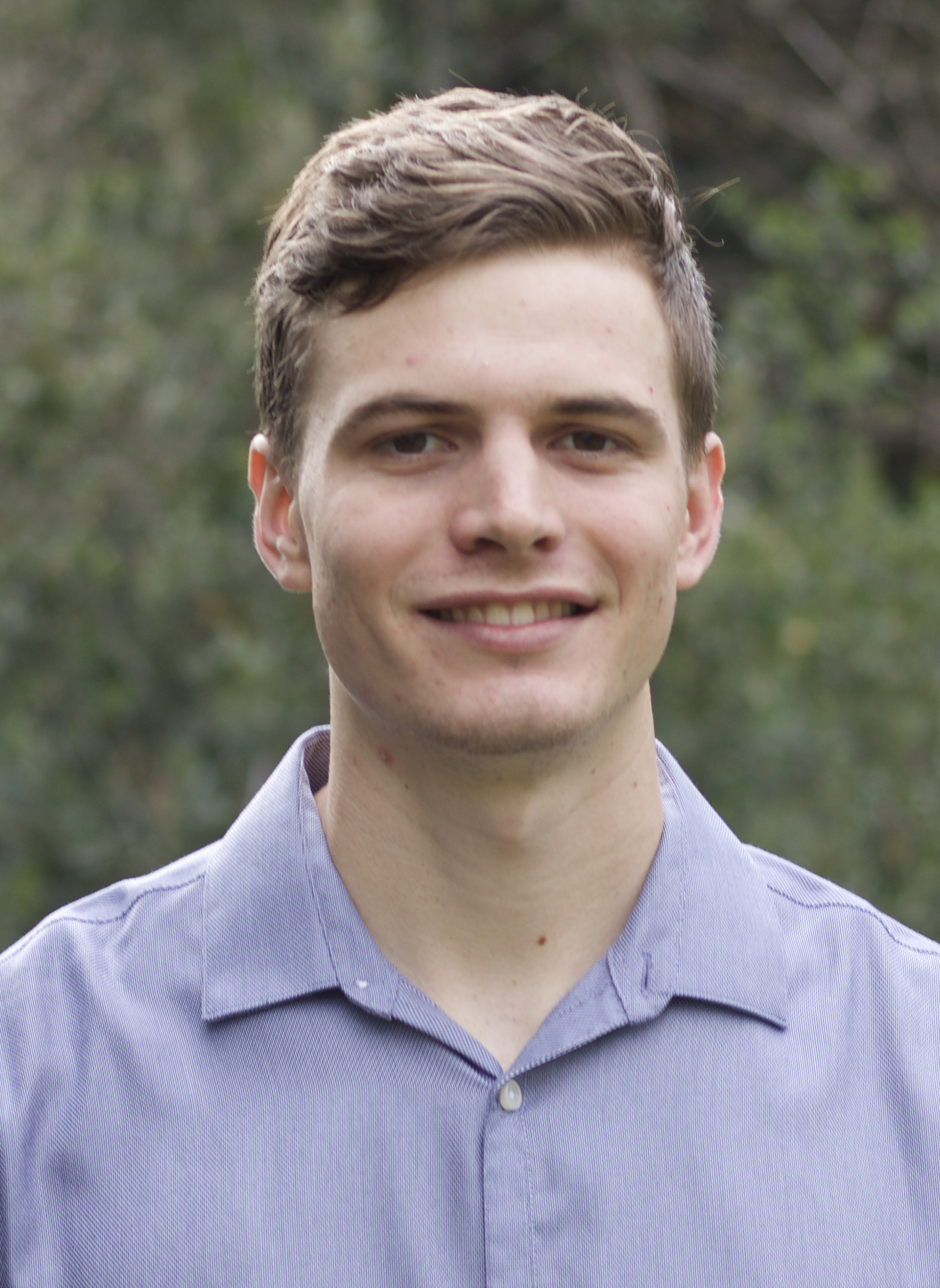}}]{Shane T. Barratt}
is currently working towards the Ph.D. degree in Electrical Engineering at Stanford University.
He received the B.S. degree in Electrical Engineering and Computer Science from the University of California, Berkeley, in 2017.
\end{IEEEbiography}
\vskip -3.5\baselineskip 
\begin{IEEEbiography}[{\includegraphics[width=1in,height=1.25in,clip,keepaspectratio]{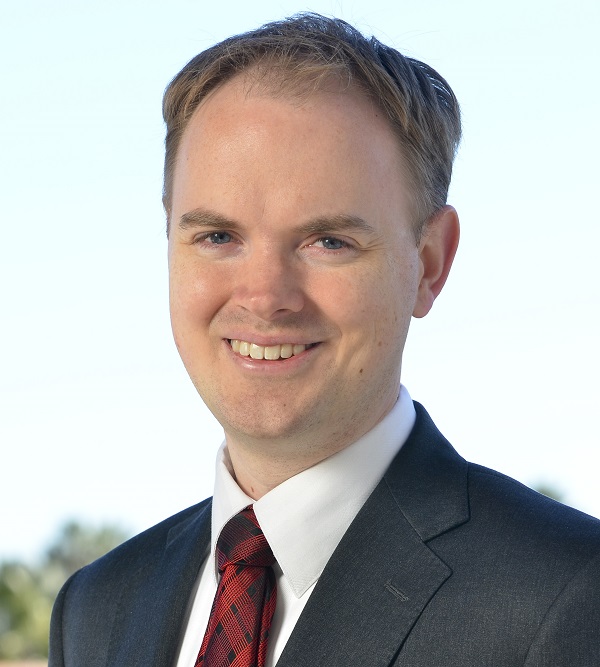}}]{Mykel J. Kochenderfer}
received the B.S. and M.S.
degrees in Computer Science from Stanford University in 2003 and the Ph.D. degree
from the University of Edinburgh in 2006.
He was with MIT Lincoln Laboratory, where he
worked on airspace modeling and aircraft collision
avoidance.
Since 2013, he has been an Assistant Professor
of Aeronautics and Astronautics at Stanford
University.
\end{IEEEbiography}
\vskip -3.5\baselineskip 
\begin{IEEEbiography}[{\includegraphics[width=1in,height=1.25in,clip,keepaspectratio]{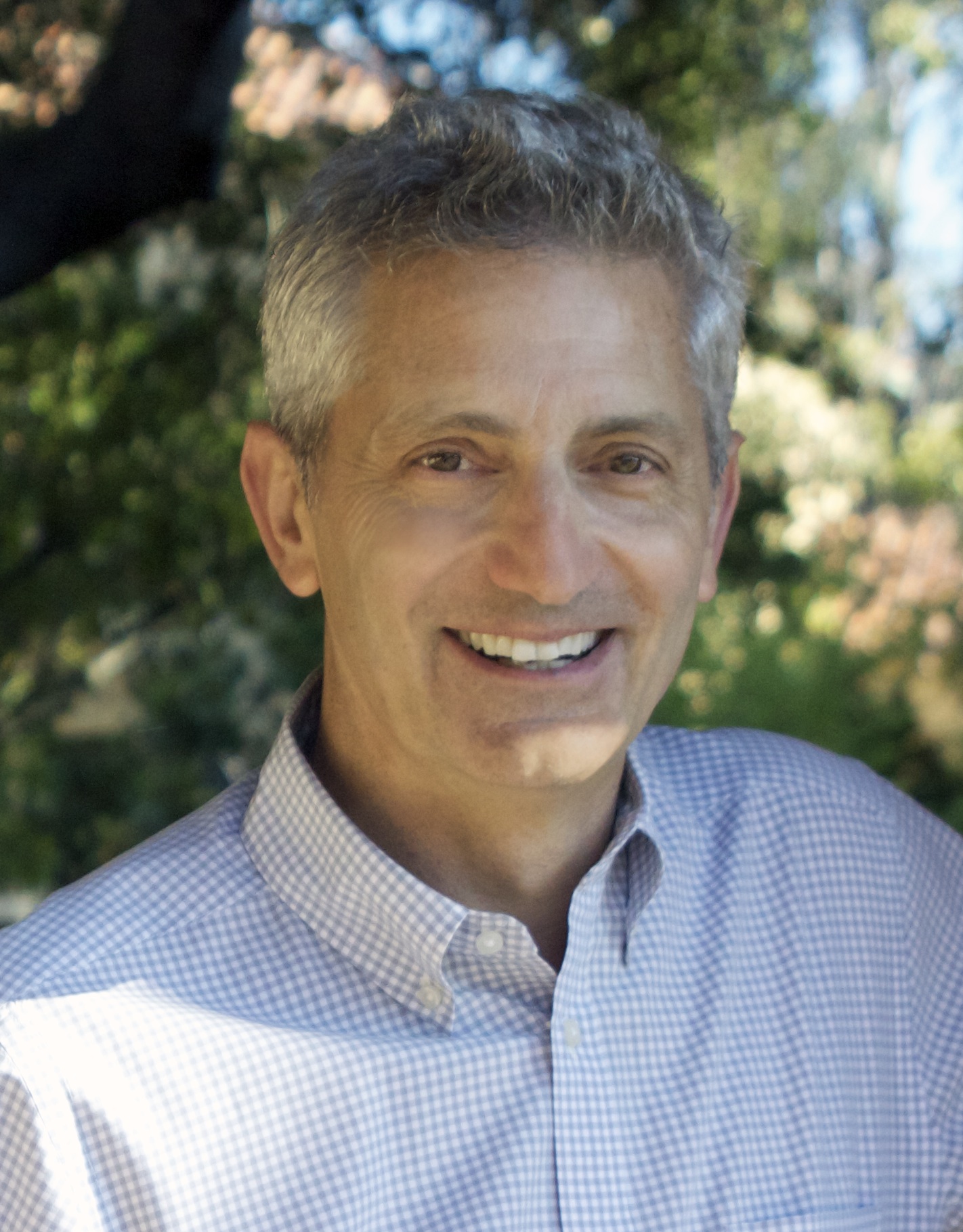}}]{Stephen P. Boyd}
is the Samsung Professor of Engineering,
and Professor of Electrical Engineering in
the Information Systems Laboratory at Stanford University, with courtesy
appointments in Computer Science and Management Science and Engineering.
He received the A.B. degree in Mathematics
from Harvard University in 1980,
and the Ph.D. in Electrical Engineering and Computer Science from the University of California, Berkeley, in 1985, and then joined the faculty at Stanford.
\end{IEEEbiography}
%



%




\end{document}